%% file: main.tex
\documentclass[sigconf]{acmart}

\usepackage{nicematrix}
\usepackage{pifont}
\newcommand{\cmark}{\ding{51}}%
\newcommand{\xmark}{\ding{55}}%
\newcommand{\bmark}{\ding{61}}
\usepackage{subfigure}
\usepackage[export]{adjustbox}
\usepackage{algorithm}
\usepackage{algorithmic}

\AtBeginDocument{%
  }

\setcopyright{acmlicensed}
\copyrightyear{2025}
\acmYear{2025}
\acmConference[KDD '25]{Proceedings of the 31st ACM
SIGKDD Conference on Knowledge Discovery and Data Mining V.2}{August 3--7,
2025}{Toronto, ON, Canada}

\acmBooktitle{Proceedings of the 31st ACM SIGKDD Conference on Knowledge
Discovery and Data Mining V.2 (KDD '25), August 3--7, 2025, Toronto, ON, Canada}

\acmDOI{10.1145/3711896.3737150}
\acmISBN{979-8-4007-1454-2/2025/08}

\begin{document}

\title{Temporal Restoration and Spatial Rewiring for Source-Free Multivariate Time Series Domain Adaptation}

\author{Peiliang Gong}
\orcid{0000-0003-2611-3145}
\affiliation{%
  \institution{Nanjing University of Aeronautics and Astronautics}
  \city{Nanjing}
  \country{China}
}
\email{plgong@nuaa.edu.cn}

\author{Yucheng Wang}
\affiliation{%
 \institution{Institute for Infocomm Research, Agency for Science Technology and Research (A*STAR)}
 \country{Singapore} \\
 \institution{Nanyang Technological University}
 \country{Singapore}
}
\email{yucheng003@e.ntu.edu.sg}

\author{Min Wu}
\affiliation{%
  \institution{Institute for Infocomm Research, Agency for Science Technology and Research (A*STAR)}
  \country{Singapore}
}
\email{wumin@i2r.a-star.edu.sg}

\author{Zhenghua Chen}
\affiliation{%
 \institution{Institute for Infocomm Research, Agency for Science Technology and Research (A*STAR)}
 \country{Singapore} \\
 \institution{Centre for Frontier AI Research, Agency for Science Technology and Research (A*STAR)}
 \country{Singapore}
}
\email{chen0832@e.ntu.edu.sg}

\author{Xiaoli Li}
\authornote{Corresponding Author}
\affiliation{%
 \institution{Institute for Infocomm Research, Agency for Science Technology and Research (A*STAR)}
 \country{Singapore} \\
 \institution{Centre for Frontier AI Research, Agency for Science Technology and Research (A*STAR)}
 \country{Singapore}
}
\email{xlli@i2r.a-star.edu.sg}

\author{Daoqiang Zhang}
\authornotemark[1]
\affiliation{%
  \institution{Nanjing University of Aeronautics and Astronautics}
  \city{Nanjing}
  \country{China}}
\email{dqzhang@nuaa.edu.cn}

\renewcommand{\shortauthors}{Peiliang Gong et al.}

\begin{abstract}
  Source-Free Domain Adaptation (SFDA) aims to adapt a pre-trained model from an annotated source domain to an unlabelled target domain without accessing the source data, thereby preserving data privacy. While existing SFDA methods have proven effective in reducing reliance on source data, they struggle to perform well on multivariate time series (MTS) due to their failure to consider the intrinsic spatial correlations inherent in MTS data. These spatial correlations are crucial for accurately representing MTS data and preserving invariant information across domains. To address this challenge, we propose Temporal Restoration and Spatial Rewiring (TERSE), a novel and concise SFDA method tailored for MTS data. Specifically, TERSE comprises a customized spatial-temporal feature encoder designed to capture the underlying spatial-temporal characteristics, coupled with both temporal restoration and spatial rewiring tasks to reinstate latent representations of the temporally masked time series and the spatially masked correlated structures. During the target adaptation phase, the target encoder is guided to produce spatially and temporally consistent features with the source domain by leveraging the source pre-trained temporal restoration and spatial rewiring networks. Therefore, TERSE can effectively model and transfer spatial-temporal dependencies across domains, facilitating implicit feature alignment. In addition, as the first approach to simultaneously consider spatial-temporal consistency in MTS-SFDA, TERSE can also be integrated as a versatile plug-and-play module into established SFDA methods. Extensive experiments on three real-world time series datasets demonstrate the effectiveness and versatility of our approach.
\end{abstract}




\maketitle

\section{Introduction}
Deep learning has achieved remarkable success in time series applications such as healthcare and human action recognition \cite{zhang2023adacket, gong2023spiking, gong2023astdf, gong2024tfac, zhang2023temporal}. However, these achievements are often contingent upon large-scale labeled datasets, which incur substantial annotation costs \cite{zhang2024diverse}. 
To alleviate this issue, unsupervised domain adaptation (UDA) has emerged as a promising alternative, aiming to adapt pre-trained models from labeled source domains to unlabelled target domains while mitigating distribution shifts \cite{UDA_Survey}.

Existing time series UDA methods predominantly rely on feature alignment or adversarial training to bridge domain gaps \cite{adatime}. While promising, these approaches often assume access to both source and target data during adaptation process. In practice, privacy concerns and resource limitations frequently lead to the unavailability of source data \cite{shot}, which hinders the broader adoption and advancement of this field. To address such limitation, a more practical setting known as source-free domain adaptation (SFDA) has been proposed, which strictly prohibits access to source data during the target adaptation phase \cite{SFDA_Survey}. Currently, various approaches have been developed for visual applications and have demonstrated satisfactory results in reducing reliance on source data \cite{GAN1,Select3,shot,aad}. Nevertheless, these methods are primarily designed for visual tasks and may not adequately account for the temporal characteristics specific to time series data. Consequently, their performance may be restricted when applied to time series contexts.

Recently, there has been a growing interest in applying SFDA to time series. One prevalent paradigm involves masking portions of the timestamps and then recovering the missing information to capture time-dependent properties \cite{MAPU, wang2024temporal}. While these methods have shown promise for time-series data, they often struggle with multivariate time-series (MTS) data. MTS data, typically collected from multiple channels that record different aspects of a system or activity simultaneously, exhibit inherent spatial correlations between channels \cite{MTGNN, wang2025survey, liu2024graph}. These spatial relationships provide valuable information that existing methods often overlook, limiting their efficacy for multivariate time-series source-free domain adaptation (MTS-SFDA). In specific, spatial information plays a pivotal role in MTS-SFDA in two key aspects. First, spatial information contributes to a richer data representation, facilitating the model to better understand the context and intricate interactions between channels. Second, spatial structures tend to be more invariant across domains compared to temporal patterns, enhancing model robustness to domain shifts. For instance, in human activity recognition, while temporal patterns of activity might vary (e.g., stride length, speed) across individuals, the spatial relationship (e.g., coordinates between limbs) usually remains relatively consistent. Therefore, we argue that considering only time-dependent properties while neglecting the modeling and migration of vital spatial correlations greatly limits the adaptability of MTS-SFDA. Accordingly, this raises a key question: \textit{How can we effectively model and adapt both spatial-temporal dependencies in MTS simultaneously without access to source data?} This premise becomes even more challenging in source-free settings.

To address the aforementioned challenges, we propose \textbf{TE}mporal \textbf{R}estoration and \textbf{S}patial r\textbf{E}wiring (TERSE), a novel SFDA approach for MTS data. Our method aims to simultaneously capture spatial-temporal dependencies within the source domain by training temporal restoration and spatial rewiring networks, which are then maintained and migrated to the target domain to guide implicit alignment and adaptation.
Specifically, we introduce two auxiliary tasks to model and transfer spatial-temporal relationships across domains. During source-domain pre-training, we first design a \textit{temporal restoration network} to capture intrinsic temporal dynamics by reconstructing temporally masked MTS signals in the latent spaces. Meanwhile, the spatial correlation structure between channels is constructed by the graph learner, and we further create a \textit{spatial rewiring network} to mine essential spatial relationships by recovering a spatially masked correlation structure of the MTS data. By doing so, the intricate spatial-temporal dependencies inherent in MTS data of the source domain can be maintained. During the target adaptation phase, the two auxiliary tasks are used to guide the feature encoder to generate output target features that can be accurately reinstated by the source-trained temporal restoration and spatial rewiring networks. This process ensures that the extracted target features are spatially and temporally consistent with the source feature distribution, thereby effectively mitigating distributional bias.

The main contributions of this work are as follows:
\begin{itemize}
    \item To our knowledge, this is the first method tailored for MTS-SFDA that explicitly considers essential channel-wise spatial relationships.
    \item We propose both temporal restoration and spatial rewiring tasks to ensure spatial-temporal consistency of sequences across domains. These can be seamlessly integrated as versatile plug-and-play modules to enhance the adaptability of established SFDA methods.
    \item We conduct extensive experiments and analyses on various real-world datasets to demonstrate the significant adaptation effectiveness and versatility of our method.
\end{itemize}

\section{Related Work}
\subsection{Source-Free Domain Adaptation}
Existing SFDA methods fall into two main categories: surrogate source generation and target model fine-tuning. Surrogate source generation attempts to create source-like data for the target domain. This involves directly selecting relevant data from the target domain or employing generative adversarial (GAN) strategies to synthesize source-style samples \cite{Select1,Select2,GAN2}. Traditional UDA strategies are then used to mitigate domain bias. In addition, target model fine-tuning approaches adapt pre-trained source models to the target domain using unlabeled target data under a self-supervised learning paradigm. Techniques like entropy regularization \cite{Entropymin1,Entropymin2}, pseudo-labels generation \cite{nrc,Pseudolabel1,Pseudolabel2}, auxiliary tasks \cite{SHOT++,MAPU,E_MAPU}, and contrast learning \cite{DAC,Contrast1} are often used to address domain shifts. While these methods have shown promise, they are primarily designed for visual tasks and may fail to consider the inherent temporal properties and spatial dependencies crucial in MTS data.
Recent video domain adaptation approaches have explored temporal consistency learning \cite{li2023source, xu2022source} and multimodal alignment \cite{huang2022relative}. However, while videos and time series share temporal properties, they differ significantly in how spatial information is represented: in video, spatial information is explicit through frame pixels, whereas in MTS, spatial relationships exist as implicit correlations between channels. This fundamental difference renders video-based adaptation methods inadequate for MTS data, which requires specialized techniques to capture the intricate spatial-temporal dependencies where channel correlations evolve continuously over time. Conversely, our method bridges this gap by simultaneously exploiting these unique spatial-temporal characteristics during adaptation, ensuring spatial-temporal consistency aligned between domains.

\subsection{Time Series Domain Adaptation}
Time series DA can be broadly categorized in two ways: feature alignment and adversarial training. Feature alignment methods focus on aligning statistical properties of features across domains \cite{ottUDA,SASA}. For instance, AdvSKM refines the maximum mean discrepancy in a hybrid spectral kernel \cite{dskn}, while RAINCOAT aligns time-frequency information to mitigate distribution bias \cite{raincoat}. Adversarial training approaches mitigate domain gaps using GAN techniques \cite{DAF,ADAST,CALDA}. For example, CoDATS employs gradient reversal layers and weak supervision for multi-source DA \cite{codats}, whereas SLARDA designs autoregressive domain discriminators that consider temporal properties \cite{SLARDA}. Recent methods also explored alternative perspectives for time series DA. For example, COTMix uses contrastive learning techniques \cite{CoTMix}, while SEA \cite{SEA} and SEA++ \cite{SEA++} address domain bias by aligning local and global sensor features and spatial interactions. POND \cite{pond} introduces prompt-based domain discrimination for multi-source time series domain adaptation, utilizing learnable prompts to capture domain-specific information and conditional modules to generate time-sensitive prompts for improved cross-domain transfer. Nevertheless, these methods often assume that source data is accessible during adaptation, which may not be feasible in real-world scenarios due to privacy or storage concerns. Recently, MAPU proposed an SFDA approach using a temporal imputation task to recover the original signal \cite{MAPU}. Although promising, it neglects the essential spatial correlations in MTS, thus restricting its adaptability.

\begin{figure*}[t]
\centering
\includegraphics[width=0.95\textwidth]{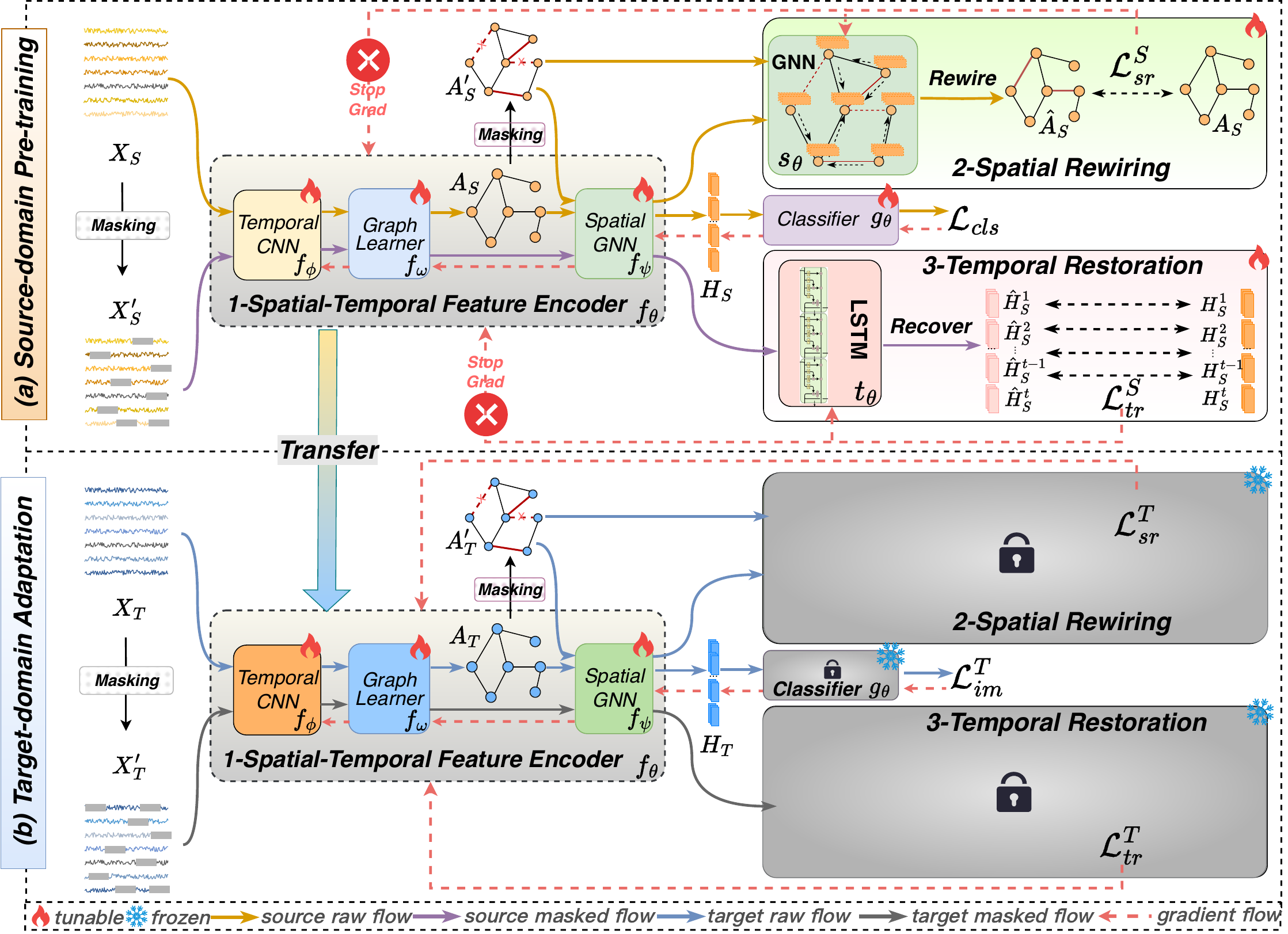}
\caption{Overview of TERSE. (a). Pre-training: The feature encoder $f_\theta$ and classifier $g_\theta$ is trained using cross-entropy loss $\mathcal{L}_{cls}$. Besides, the spatial rewiring network $s_\theta$ and temporal restoration network $t_\theta$ are trained by $\mathcal{L}_{sr}^{S}$ and $\mathcal{L}_{tr}^{S}$. Gradient stopping is applied here to ensure that auxiliary tasks do not negatively affect the pre-training performance. (b). Adaptation: The target feature encoder $f_\theta$ is jointly fine-tuned with both information maximum loss $\mathcal{L}_{im}^{T}$ and our proposed spatial rewiring loss $\mathcal{L}_{sr}^{T}$ and temporal restoration loss $\mathcal{L}_{tr}^{T}$ while ensuring the spatial-temporal consistency adaptation with source domain.}
\label{fig_framework}
\end{figure*}
\vspace{-0.2cm}

\section{Methodology}
\subsection{Problem Definition}
We follow established vendor-client paradigm, which reflects real-world scenarios where vendors protect source data while providing adaptable solutions to clients. This paradigm, adopted in many recent works \cite{vc1,vc3,MAPU}, enables pretraining customization while maintaining strict source data privacy during target adaptation. 
Given a labeled source domain $\mathcal{D}_S = \{{X}_S^i, {y}_S^i\}_{i=1}^{n_S} $ with $n_S$ samples and an unlabeled target domain $\mathcal{D}_T= \{{X}_T^i\}_{i=1}^{n_T} $ with $n_T$ samples, both ${X}_S$ and ${X}_T$ represent MTS data with $N$ channels and $L $ sequence lengths, and ${y}_S$ denotes the labels of $X_S$. We assume a shift in the marginal distributions between domains (i.e., $P(X_S) \neq P(X_T)$), while the conditional distributions are presumed to remain similar (i.e., $P(y_S|X_S) \approx P(y_T|X_T)$).

\subsection{Overview}
We propose a novel method to address the challenge of MTS-SFDA, considering both spatial-temporal dependencies across domains, as in Fig. \ref{fig_framework}. Our method comprises two stages: (1) During source-domain pre-training, given MTS data, we employ a dual-pronged approach. First, a graph learner extracts inter-channel interactions, which are then spatially masked. Using this masked graph structure, we train a spatial GNN as a \textit{spatial rewiring network} to capture the source spatial characteristics. Besides, we apply a masking strategy to the original signals and train an encoder network as a \textit{temporal restoration network} to capture the source domain's temporal dynamics. (2) During target-domain adaptation, we utilize source pre-trained \textit{temporal restoration and spatial rewiring networks} to guide the target spatial-temporal feature encoder. This encoder generates target features that are temporally consistent and spatially correlated with the source domain, thereby implicitly enhancing feature alignment across domains and facilitating adaptation capabilities. The details of our method are described below.

\subsection{Source-Domain Pre-Training}
In this stage, we train three key components, i.e., the feature encoder, spatial rewiring, and temporal restoration. In specific, the feature encoder aims to extract underlying spatial-temporal characteristics from the MTS data. Meanwhile, both the spatial rewiring task and the temporal restoration task are designed to capture the spatial relationships and the temporal dynamics in the source domain.

\subsubsection{Source Feature Encoder and Classifier Optimization}
We create a deep architecture to learn spatial-temporal characteristics of MTS data. It includes two parts: a spatial-temporal feature encoder $f_\theta: \mathcal{X}_S \rightarrow \mathcal{H}_S$ and a linear classifier $g_\theta:\mathcal{H}_S \rightarrow \mathcal{Y}_S$. The spatial-temporal feature encoder comprises three parts, a temporal CNN, a graph learner, and a spatial GNN, details of which are given below.

\begin{itemize}
    \item \textbf{\textit{Temporal Convolutional Neural Network (CNN)}} This module employs a 3-layer 1-D CNN, denoted as $f_\phi:\mathcal{X}_S \rightarrow \mathcal{Z}_S$, to extract temporal features from each channel signal. Each convolutional layer is followed by a rectified linear unit (ReLU) non-linear activation function, a Batch Normalization layer, and a Max Pooling layer.
    \item \textbf{\textit{Graph Learner}} The extracted temporal features are then fed into a graph learner to capture the spatial correlations between channels, denoted as $f_\omega:Z_S \rightarrow A$. For simplicity, the graph learner utilizes the inner product operation between the features. The resulting correlation matrix is computed as follows:
    \begin{equation}
        A =\sigma(\tilde{Z}_S \cdot \tilde{Z}^{\prime}_{S}), \quad \tilde{Z}_S=\frac{Z_S}{{\Vert Z_S \Vert}_2} 
    \end{equation}
    where $\sigma$ is a ReLU activation function, ${\Vert\cdot\Vert}_2$ is $L_2$ norm.
    \item \textbf{\textit{Spatial Graph Neural Network (GNN)}} Both the temporal features and the learned spatial correlation matrix are fed into a 1-layer GNN \cite{GNN}. This GNN consists of two key phases: propagation and update. During propagation, features from neighboring nodes are aggregated and passed to the central node. Subsequently, the aggregated features are updated using a linear transformation layer, as formulated below:
    \begin{equation}
        f_\psi(Z_S, A)=H_S=\sigma(\tilde{D}^{-\frac{1}{2}} A \tilde{D}^{-\frac{1}{2}}Z_SW) 
    \end{equation}
    where $\tilde{D}_{i,i}=\sum_{j}A_{i,j}$, $W$ is a learnable weight matrix, $\sigma$ denotes the PReLU \cite{PReLU} activation function.
\end{itemize}

We optimize the source feature encoder and classifier by minimizing the following cross-entropy loss incorporating label smoothing technique \cite{label_smoothing}.
\begin{align}
    \min_{f_{\theta}, g_{\theta}}\mathcal{L}_{cls} 
    &= - \sum_{i=1}^{n_S} \sum_{k=1}^{K} \hat{y}_{k} \log {\delta_{k}(g_{\theta}(f_{\theta}(X_{S}^{i})))},
\end{align}
where $\delta_{k}(\cdot)$ denotes the softmax function, $\hat{y}_{k}$ represents the $k^{th}$ element of the smooth label $\hat{y}_{k} = (1-\eta)y_k + \eta / K$, $\eta$ is the smoothing coefficient which is empirically set to $0.1$,  and $y_k$ is the one-hot encoding of label.

\subsubsection{Source Temporal Dynamics Capturing}
To capture the intrinsic temporal dependencies within the source domain, we design a temporal restoration auxiliary task. This task enables the model to capture temporal relationships by reconstructing temporally masked MTS signals in the latent spaces. Specifically, given the original signal $X_S$, we create a temporally masked version $X_{S}^{\prime}$ using a temporal masking strategy. This strategy divides the input signal into non-overlapping segments and randomly masks a portion of them by setting the values to zero. We then employ the designed spatial-temporal feature encoder to extract latent representations $H_S$ and $H_{S}^{\prime}$ for both original and masked signals.

The temporal restoration task is then introduced by employing an additional \textit{temporal restoration network} $t_\theta$. This network consists of a single-layer LSTM that aims to map the masked latent representation $H_{S}^{\prime}$ back to the original version, which can be denoted as $t_\theta: H_{S}^{\prime} \rightarrow \hat{H}_{S}$. The \textit{temporal restoration network} is optimized by minimizing the mean squared error (MSE) between the reconstructed latent representation $\hat{H}_{S}$ and the original latent representation $H_S$,
\begin{align}
    \min_{t_{\theta}}\mathcal{L}_{tr}^{S} 
    &= \frac{1}{n_S}\sum_{i=1}^{n_S} \left\|H_{S}^{i} - \hat{H}_{S}^{i} \right\|_2^2 ,   
\end{align}
where $H_S^i = f_\theta(X_S^i)$ is representations of the original signal, $\hat{H}_S^i = t_\theta(f_\theta(X_S^{\prime i}))$ is restored representation by the \textit{temporal restoration network}. By learning to restore the missing information using the temporal context surrounding the masked segments, the restoration network is encouraged to exploit the temporal dynamics inherent within the sequence.

\subsubsection{Source Spatial Correlation Capturing}
To exploit the spatial correlations between channels in the source domain, we design a spatial rewiring auxiliary task. This task encourages the model to capture essential spatial relationships by reconstructing a spatially masked correlation structure of the MTS. Given the input $X_S$, we first obtain the original spatial correlation structure $A_S$ using the graph learner. A masking strategy then creates a spatially masked version $A_S^{\prime}$ by randomly removing a portion of the edges and setting their weight to zero. We then feed the extracted features $Z_S$ from the temporal CNN and the masked graph $A_S^{\prime}$ into the spatial GNN to obtain a corresponding latent representation $H_S^{\prime \prime}$.

The spatial rewiring task is then introduced through an additional \textit{spatial rewiring network} $s_\theta:\{H_S^{\prime \prime}, A_S^{\prime}\} \rightarrow \hat{A}_S$. This network consists of a single-layer GNN. It first maps the representation $H_S^{\prime \prime}$ learned from the masked graph to an embedding $\hat{Z}_S$. Then, it utilizes pairwise similarity within this embedding to rewire the masked-out edges, as follows,
\begin{align}
    \hat{A}_S= \frac{\hat{Z}_S}{\Vert \hat{Z}_S \Vert_2} \cdot (\frac{\hat{Z}_S}{\Vert \hat{Z}_S \Vert_2})^\prime, \hat{Z}_S=\sigma(\hat{D}^{-\frac{1}{2}} A_S^{\prime} \hat{D}^{-\frac{1}{2}} H_S^{\prime \prime} W^{\prime}),
\end{align}
where $\hat{D}_{i,i}=\sum_{j} {A_S^{\prime}}_{i,j}, W^\prime$ is a learnable weight matrix, $\sigma$ denotes the PReLU activation function \cite{PReLU}. The network is trained by minimizing the MSE between the rewired spatial correlation structure $\hat{A}_S$ and the original structure $A_S$,
\begin{align}
    \min_{s_{\theta}}\mathcal{L}_{sr}^{S} 
    &= \frac{1}{n_S}\sum_{i=1}^{n_S} \left\|A_{S}^{i} - \hat{A}_{S}^{i} \right\|_2^2 ,
\end{align}
here, $\hat{A}_S^i=s_\theta(A_S^{\prime i}, f_\psi(A_S^{\prime i}, f_\phi(X_S^i)))$, $A_S^i=f_\omega(f_\phi(X_S^i))$.

Finally, the source-domain pre-training is achieved by jointly minimizing the following combined loss function:
\begin{align}
    \mathcal{L}_{src}^S=\mathcal{L}_{cls} + \mathcal{L}_{tr}^{S} + \mathcal{L}_{sr}^S.
\end{align}

Notably, the designed auxiliary tasks aim to capture the spatial-temporal dependencies in the source domain for target adaptation. To prevent them from hindering the pre-training performance of the feature encoder $f_\theta$, we employ a gradient-stopping strategy during the backpropagation. This ensures that the gradients from the auxiliary losses do not affect the update of the feature encoder's parameters.

\subsection{Target-Domain Adaptation}
This stage aims to adapt the pre-trained encoder to the target domain without access to source data. To maintain the spatial correlations and temporal dependencies learned in the source domain, we employ the temporal restoration and spatial rewiring tasks trained in the pre-training stage. These tasks guide encoder $f_\theta$ in regularizing the extracted target features, ensuring that they are aligned with the spatial-temporal characteristics of the source domain.

\subsubsection{Temporal Adaptation with Feature Restoration}
To train the feature encoder to produce target features temporally aligned with the source domain, we leverage pre-trained \textit{temporal restoration network} $t_\theta$ as a guiding force during feature encoder optimization. First, the spatial-temporal encoder extracts representations $H_T$ and $H_T^\prime$ for the original $X_T$ and temporally masked target domain sequences $X_T^\prime$, respectively. Then, a frozen source pre-trained \textit{temporal restoration network} $t_\theta$ operates in latent spaces, attempting to reconstruct the original sequence's representation $H_T$ from the masked one $H_T^\prime$. Due to domain bias, the source-trained restoration network might not perfectly restore the target domain information. Therefore, we adjust the target feature encoder by minimizing the temporal restoration loss. This, in turn, encourages the target features to retain information that aligns with the source temporal characteristics, allowing for successful reconstruction by the restoration network.
\begin{align}
\label{min_loss_tr_T}
    \min_{f_{\theta}}\mathcal{L}_{tr}^{T} 
    &= \frac{1}{n_T}\sum_{i=1}^{n_T} \left\|H_{T}^{i} - \hat{H}_{T}^{i} \right\|_2^2 ,
\end{align}
where $H_T^i = f_\theta(X_T^i)$ is the latent representation of the original signal, $\hat{H}_T^i = t_\theta(f_\theta(X_T^{\prime i}))$ is the restored latent representation by the frozen temporal restoration network.

\subsubsection{Spatial Adaptation with Graph Rewiring}
We further use pre-trained \textit{spatial rewiring network} $s_\theta$ to guide the target feature encoder's optimization towards generating output features with spatial correlations aligned with the source domain. The target graph learner first extracts the original spatial correlation structure $A_T$ and creates a masked version $A_T^\prime$ using a masking strategy. The spatial GNN then processes the masked structure $A_T^\prime$ to obtain the corresponding representation $H_{T}^{\prime \prime}$. Subsequently, the frozen source pre-trained \textit{spatial rewiring network} $s_\theta$ attempts to reconstruct original spatial correlations from the masked data. To encourage the target feature encoder to produce features that enable better reconstruction by the \textit{spatial rewiring network} $s_\theta$, we minimize the following rewiring loss during model optimization,
\begin{align}
\label{min_loss_sr_T}
    \min_{f_{\theta}}\mathcal{L}_{sr}^{T} 
    &= \frac{1}{n_T}\sum_{i=1}^{n_T} \left\|A_{T}^{i} - \hat{A}_{T}^{i} \right\|_2^2 ,
\end{align}
here, $\hat{A}_T^i=s_\theta(A_T^{\prime i}, f_\psi(A_T^{\prime i}, f_\phi(X_T^i)))$, $A_T^i=f_\omega(f_\phi(X_T^i))$.

Further, to enhance target adaptation, we also introduce an information maximization loss \cite{shot}, which encourages the model to generate individually deterministic yet globally diverse target outputs, as follows,
\begin{align}
    \mathcal{L}_{im}^T &= -\frac{1}{n_T}\sum_{i=1}^{n_T}\sum_{k=1}^{K}p_k^i \log p_k^i + \sum_{k=1}^{K}\hat{p}_k \log \hat{p}_k,
\end{align}
where $p_k^i=\delta_K(g_\theta(f_\theta(X_T^i)))$, $\hat{p}_k=\frac{1}{n_T}\sum_{i=1}^{n_T} p_k^i$, and $\delta_{k}(\cdot)$ denotes the softmax function.
Finally, the overall target domain adaptation loss is then formulated as follows:
\begin{align}
    \mathcal{L}_{trg}^T=\mathcal{L}_{im}^T + \alpha \mathcal{L}_{tr}^{T} + \beta \mathcal{L}_{sr}^T,
\end{align}
where $\alpha$ and $\beta$ are used for controlling the relative weight between these loss components.

Algorithm \ref{alg_terse} outlines the adaptation process involving temporal restoration and spatial rewiring. For temporal adaptation, an input target signal, $X_T$, is subjected to temporal masking to create a masked version, $X_T^\prime$. Both the original and masked signals are then encoded by the pre-trained spatial-temporal feature encoder, $f_\theta$, yielding latent representations $H_T$ and $H_T^\prime$, respectively. Subsequently, the \textit{temporal restoration network}, $t_\theta$, is trained to reconstruct $H_T$ from $H_T^\prime$. For spatial adaptation, the original spatial correlation structure, $A_T$, is obtained using a graph learner. A masked version, $A_T^\prime$, is then generated. The spatial GNN module, $f_\psi$, extracts latent features from the masked correlation structure. The \textit{spatial rewiring network}, $s_\theta$, is trained to reconstruct the original spatial correlation structure, $A_T$, from the masked version. To align the target encoder with the source domain, the model is optimized to minimize the reconstruction errors of both the temporal and spatial restoration tasks, as defined in Eqs. (\ref{min_loss_tr_T}) and (\ref{min_loss_sr_T}).

\subsection{Plug-and-Play Auxiliary Tasks for SFDA}
Notably, this work introduces both temporal restoration and spatial rewiring auxiliary tasks, which can function as universal plug-and-play modules, seamlessly integrated into established SFDA approaches. Specifically, during source domain pre-training, the tasks act as guides, helping the model extract inherent temporal dynamics and spatial correlations within the source data. Whereas, in the target domain adaptation phase, the combined losses from both tasks, along with the general SFDA loss, are used to jointly optimize the target feature encoder, enhancing its spatial-temporal adaptation capabilities. A detailed exploration of the model's versatility is provided in the experimental section.

\begin{algorithm}[tb]
\caption{Adaptation with both Temporal Restoration and Spatial Rewiring}
\label{alg_terse}
\textbf{Input}: Target samples $X_T$, source pre-trained spatial-temporal feature encoder $f_\theta$ (temporal CNN $f_\phi$, graph learner $f_\omega$, and spatial GNN $f_\psi$), classifier $g_\theta$, temporal restoration network $t_\theta$, and spatial rewiring network $s_\theta$\\
\textbf{Output}: Adapted encoder $f_\theta$
\begin{algorithmic}[1] 
\STATE \textcolor{gray}{\textbf{\textit{$\rightarrow$ Temporal Restoration:}}}
\STATE $X_T^\prime=t\_masking(X_T)$ \textcolor{gray}{\COMMENT{Temporally masked signals}}
\STATE $H_T=f_\theta(X_T)$ \textcolor{gray}{\COMMENT{Latent features of raw signals}}
\STATE $H_T^\prime=f_\theta(X_T^\prime)$ \textcolor{gray}{\COMMENT{Latent features of masked signals}}
\STATE $\hat{H}_T=t_\theta(H_T^\prime)$ \textcolor{gray}{\COMMENT{Restore masked latent features}}
\STATE \textcolor{gray}{\textbf{\textit{$\rightarrow$ Spatial Rewiring:}}}
\STATE $A_T=f_\omega(f_\phi(X_T))$ \textcolor{gray}{\COMMENT{Spatial correlation structure}}
\STATE $A_T^\prime=s\_masking(A_T)$ \textcolor{gray}{\COMMENT{Spatially masked correlation}}
\STATE $H_T^{\prime \prime}=f_\psi(f_\phi(X_T), A_T^\prime)$ \textcolor{gray}{\COMMENT{Latent features with masked correlation}}
\STATE $\hat{A}_T=s_\theta(H_T^{\prime \prime}, A_T^\prime)$ \textcolor{gray}{\COMMENT{Rewire masked correlation}}
\STATE \textcolor{gray}{\textbf{\textit{$\rightarrow$ Optimization:}}}
\STATE $\min_{f_{\theta}}\mathcal{L}_{tr}^{T}(H_T, \hat{H}_T)+\mathcal{L}_{sr}^{T}(A_T, \hat{A}_T)$ \textcolor{gray}{\COMMENT{Update feature encoder $f_\theta$ with Eqs. (8) and (9).}}
\end{algorithmic}
\end{algorithm}

\section{Experiments}
\subsection{Datasets and Settings}
\subsubsection{Datasets}
We evaluate model performance on three widely used real-world time series datasets, namely human activity recognition (UCIHAR) \cite{uciHAR_dataset}, Wireless Sensor Data Mining (WISDM) \cite{wisdm_dataset}, and sleep stage classification (SSC) \cite{sleepEDF_dataset}. The details of each dataset are given in Appendix \ref{appendix_datasets}.

\subsubsection{Baselines}
We compare our method with established UDA methods and cutting-edge SFDA methods re-implemented within our framework using the same backbone and training configurations. We also provide source-only and target-supervised performance to establish lower/upper bounds. The description of each baseline is shown in Appendix \ref{appendix_baselines}. 
To ensure robust evaluation under potential data imbalance, we employ the macro F1-score (MF1) metric. We report the mean and standard deviation of the metric across three runs for each cross-domain scenario.

\subsubsection{Implementation Details}
To ensure a meticulously controlled comparison, all methods employ the proposed spatial-temporal feature encoder as the backbone network to extract spatial-temporal features. Specifically, the temporal CNN includes a 3-layer 1-D CNN with filter sizes of 64, 128, and 128 respectively. The spatial GNN includes a single-layer GNN with an output dimension of 256 features. Besides, we employ a consistent masking ratio of 1/8 for temporal masking and 0.5 for spatial masking across all datasets. In accordance with the AdaTime framework \cite{adatime}, we train all models for 40 epochs with a batch size of 32. The learning rate is set to 1e-3 for UCIHAR and 3e-3 for both SSC and WISDM datasets. For each algorithm/dataset combination, we conduct an extensive random hyper-parameter search to identify optimal configurations with fifty hyper-parameter combinations. The hyper-parameters are picked by a uniform sampling from a range of predefined values. We built our model using Pytorch and trained it on an NVIDIA GeForce RTX 2080Ti GPU. This meticulous approach guarantees that any observed performance discrepancies can be solely attributed to the underlying algorithmic design, as opposed to variations in network architecture or training configurations.

\input{Tables/main/ucihar_main_results}
\input{Tables/main/wisdm_main_results}
\input{Tables/main/eeg_main_results}

\subsection{Comparative Experiments}
Tables \ref{table:har}-\ref{table:ssc} detail the results for five cross-domain scenarios within each dataset, along with the average performance across all scenarios. For clarity, the algorithms are categorized into three groups: lower/upper bounds are denoted by \bmark, traditional UDA methods by \xmark, and SFDA methods by \cmark.

\subsubsection{Quantative Results on UCIHAR Dataset}
Table \ref{table:har} presents a comprehensive evaluation of our method across five cross-subject scenarios on the UCIHAR dataset. Our approach achieves superior performance with a mean F1-score of 97.50\%, demonstrating state-of-the-art results in three out of five scenarios. Notably, our method surpasses the second-best SFDA approach by 1\%, while also outperforming UDA methods that have the advantage of accessing source data. The strong performance of SFDA methods, comparable to or exceeding UDA approaches, can be attributed to their two-stage training strategy (pretraining and adaptation) that enables focused optimization on the target domain. Our method's dual adaptation mechanism, incorporating both temporal and spatial components, proves particularly effective, surpassing the best UDA method (i.e., CDAN) by 3.13\% and demonstrating a substantial improvement of 16\% over the source-only baseline. These results provide compelling evidence for the effectiveness of our proposed approach in addressing cross-domain adaptation challenges in HAR tasks.

\subsubsection{Quantative Results on WISDM Dataset}
Table \ref{table:wisdm} presents a comprehensive comparison of our method against existing approaches across five cross-domain scenarios on the WISDM dataset. Our approach demonstrates superior overall performance, achieving a mean F1-score of 73.84\%. While UDA methods exhibit advantages in certain scenarios due to their access to source data, our source-free approach shows remarkable effectiveness, outperforming baseline approaches in several key aspects. Compared to other SFDA methods, we achieve superior performance in three out of five scenarios, surpassing the second-best SFDA method (i.e., MAPU) by a substantial margin of 4.35\%. Despite the inherent disadvantage of not accessing source data, our method exceeds the best-performing UDA approach (CoDATs) by 4.57\%, and most notably, shows a significant improvement of over 20\% compared to the source-only baseline. The WISDM dataset presents unique challenges due to its inherent class imbalance within individual subjects' data; however, our method's robust performance can be attributed to the synergistic effect of our dual adaptation mechanism - temporal restoration and spatial rewiring. These complementary strategies enhance the model's adaptability by effectively addressing both temporal dependencies and spatial feature distributions, contributing to more robust cross-domain generalization and demonstrating the method's capability to handle complex domain shifts while maintaining classification accuracy under challenging conditions.

\subsubsection{Quantative Results on SSC Dataset}
Table \ref{table:ssc} presents a comprehensive evaluation of our method on sleep stage classification tasks across five cross-domain scenarios. Our approach achieves state-of-the-art performance with a mean F1-score of 68.40\%, outperforming both SFDA and conventional UDA methods by margins of 2.06\% and 0.62\%, respectively. The superior performance is particularly evident in scenarios 16$\rightarrow$1 and 7$\rightarrow$18, where our method demonstrates improvements of 2.02\% and 3.04\% over existing SFDA approaches. However, we observe negative transfer in certain scenarios compared to the source-only baseline, particularly in scenario 0$\rightarrow$11, where performance decreases by 4.22\%. This phenomenon can be attributed to the significant inter-individual variations in EEG signals and severe sleep stage imbalances, which pose substantial challenges for adaptation. Besides, while feature clustering-based methods like SF(DA)$^2$ show limitations on the inherently imbalanced SSC dataset, our method's dual adaptation mechanism generally mitigates these challenges through integrated temporal restoration and spatial rewiring strategies. The overall improvement across diverse scenarios, despite these inherent difficulties, demonstrates our method's effectiveness in handling the complex domain adaptation challenges in SSC tasks.

\subsection{Ablation Study}
We conduct ablation studies to evaluate the key components of our method. First, we analyze the contribution of each auxiliary task to the overall adaptation performance. Further, we also assess the efficacy of spatial-temporal feature encoder, as in Appendix \ref{appendix_feature_encoder}.

\subsubsection{Impact of spatial rewiring and temporal restoration}
We compare our method with three variants. Table \ref{tab:ablation_tasks} presents the average performance across five cross-domain scenarios on three datasets. The results demonstrate that omitting both temporal restoration and spatial rewiring tasks leads to the most significant performance decline. This underscores the overall importance of information recovery for target adaptation. Besides, removing either temporal restoration or spatial rewiring alone results in varying degrees of performance degradation. Notably, the SSC and WISDM datasets exhibit a more substantial drop in performance when spatial rewiring tasks are absent. This suggests that modeling and adapting spatial correlations plays an even more crucial role.

\input{Tables/abalation/masking_abalation}

\subsection{Model Analysis}
In this section, we present comprehensive analyses to validate the effectiveness and robustness of our proposed approach. We examine the model's versatility across different domains, investigate parameter sensitivity to various hyperparameter settings, and evaluate the impact of different mask ratios on adaptation performance. Furthermore, we provide feature visualization analysis to offer insights into the learned representations and adaptation mechanisms. 

\subsubsection{Versatility Analysis}
To assess the versatility of our temporal restoration and spatial rewiring approach, we integrate it with established SFDA methods among three datasets. Table \ref{tab:versatility} presents the average macro F1 score across five cross-domain scenarios. The results show that incorporating our auxiliary tasks significantly improves performance in most cases. For instance, on the UCIHAR dataset, SHOT and NRC achieve gains exceeding 1\%. On the SSC dataset, SHOT and AaD improve by 1\% and 2\%. Notably, all methods exhibit substantial improvements of at least 4\% on WISDM. These findings highlight the efficacy of our spatial-temporal adaptation strategy in enhancing the performance of existing SFDA methods.

\input{Tables/versatility/versatility_analysis}

\subsubsection{Parameter Sensitivty Analysis}
\begin{figure}[t]
\centering
\setlength{\subfigcapskip}{2pt} 
\begin{tabular}{c}
\hspace{-5mm}
\subfigure[Temporal restoration]{\includegraphics[width=0.48\linewidth]{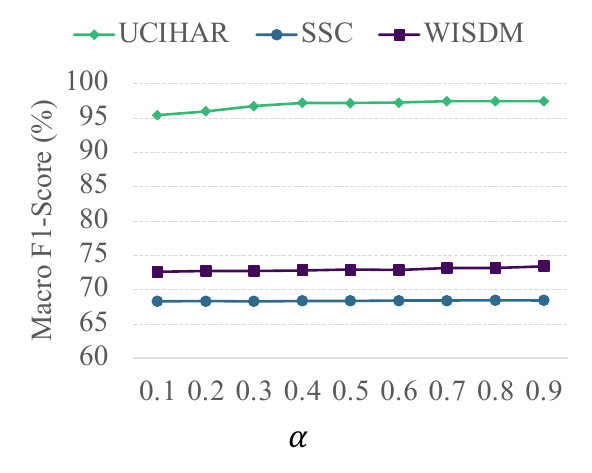} \label{fig_feat_recover_param}}
\hspace{-3mm} 
\subfigure[Spatial rewiring]{\includegraphics[width=0.48\linewidth]{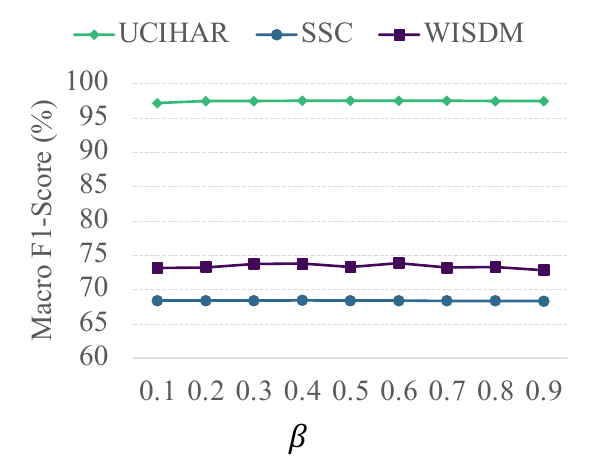} \label{fig_graph_recover_param}}

\end{tabular}
\caption{Effect of varying weight for the model parameters.}
\label{fig_recover_param}
\end{figure}

This section investigates the sensitivity of model performance to the relative weights assigned to the temporal restoration and spatial rewiring components ($\alpha$ and $\beta$, respectively). Figure \ref{fig_recover_param} presents the mean F1-score across three datasets for various weight combinations. The results indicate that the model exhibits relative stability across a range of parameter values. Notably, on the UCIHAR dataset, performance for the $\alpha$ parameter increases and then plateaus with increasing weight, demonstrating minimal overall fluctuation. The SSC and WISDM datasets show smooth performance curves across different $\alpha$ values. Similarly, $\beta$ demonstrates minimal impact on performance across all datasets. These observations suggest that the model is relatively insensitive to variations in the weights of the temporal restoration and spatial rewiring tasks.

\subsubsection{Visualization Analysis}
\begin{figure}[t]
\centering
\begin{tabular}{c}
\subfigure[Source-only]{\includegraphics[width=0.48\linewidth]{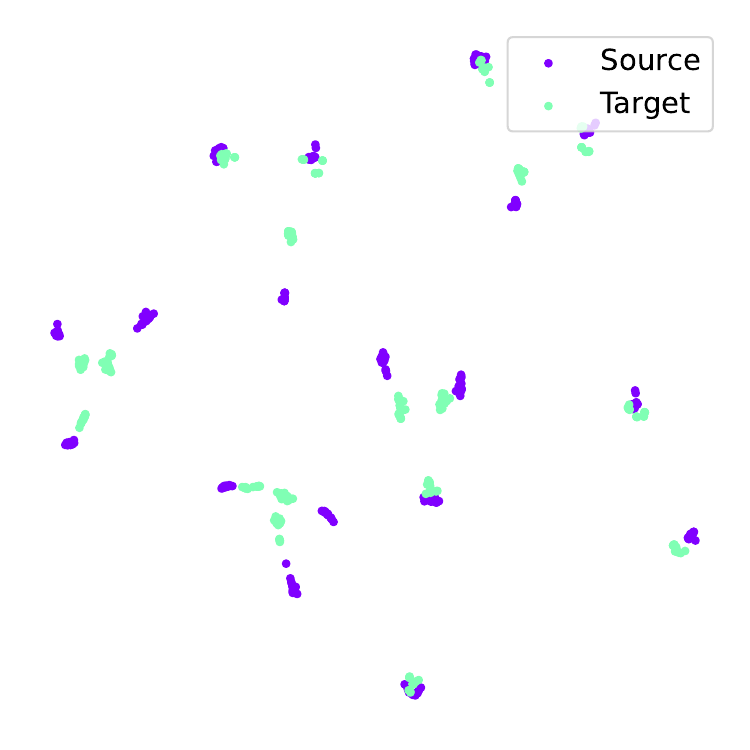} \label{fig_tsne_noadapt}}
\subfigure[Ours]{\includegraphics[width=0.48\linewidth]{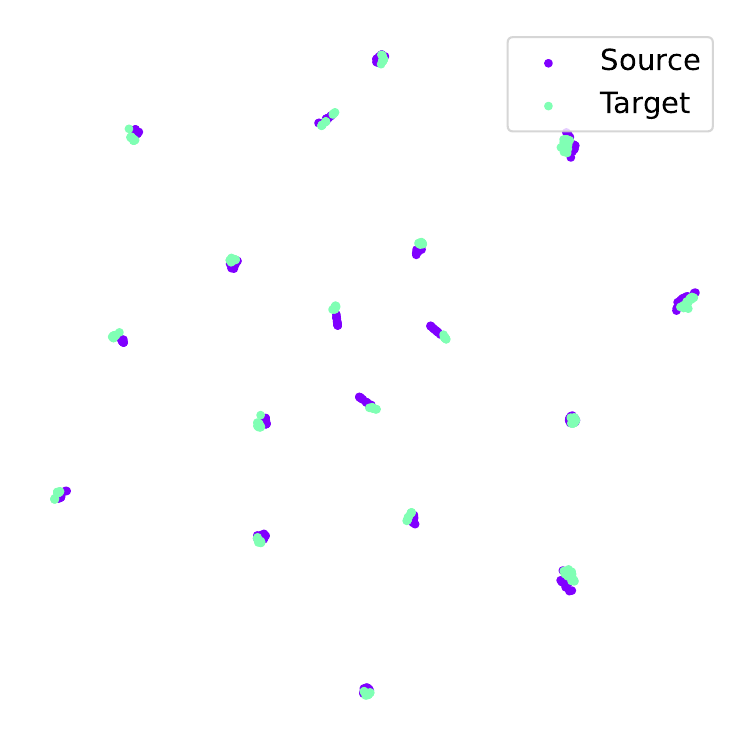} \label{fig_tsne_ours}}

\end{tabular}
\caption{The t-SNE feature visualizations of source and target domains on the UCIHAR dataset.}
\label{fig_tsne}
\end{figure}

To offer visual validation of model's adaptation capabilities, here we employ the t-SNE visualization technique \cite{tSNE}. Figure \ref{fig_tsne} depicts the feature distributions of the extracted features for both the source-only and our proposed method, respectively. Different colors represent the source and target domains. The results demonstrate that our TERSE achieves significantly better alignment between source and target domain features compared to the source-only approach. This improvement can be attributed to the model's ability to leverage both temporal and spatial adaptability.

\subsubsection{Impact of Masking Level}
\begin{figure}[t]
\centering
\setlength{\subfigcapskip}{10pt} 
\begin{tabular}{c}
\hspace{-3mm}
\subfigure[On the UCIHAR]{\includegraphics[width=0.48\linewidth]{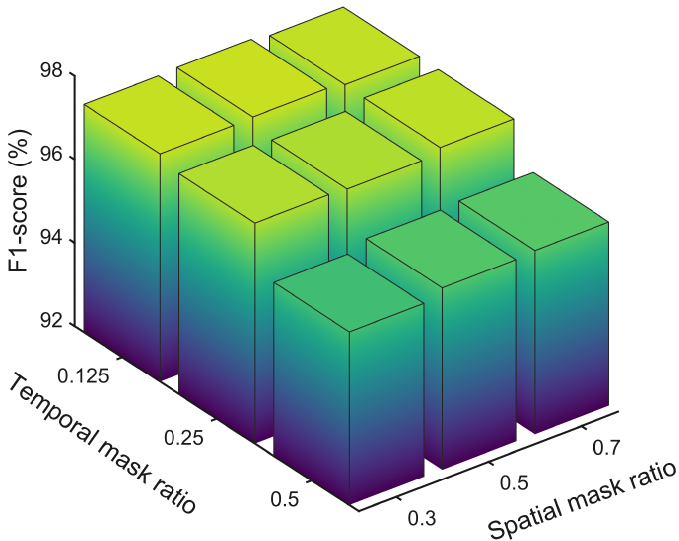} \label{fig_masking_ratio_har}} 
\hspace{-3mm}
\subfigure[On the SSC]{\includegraphics[width=0.48\linewidth]{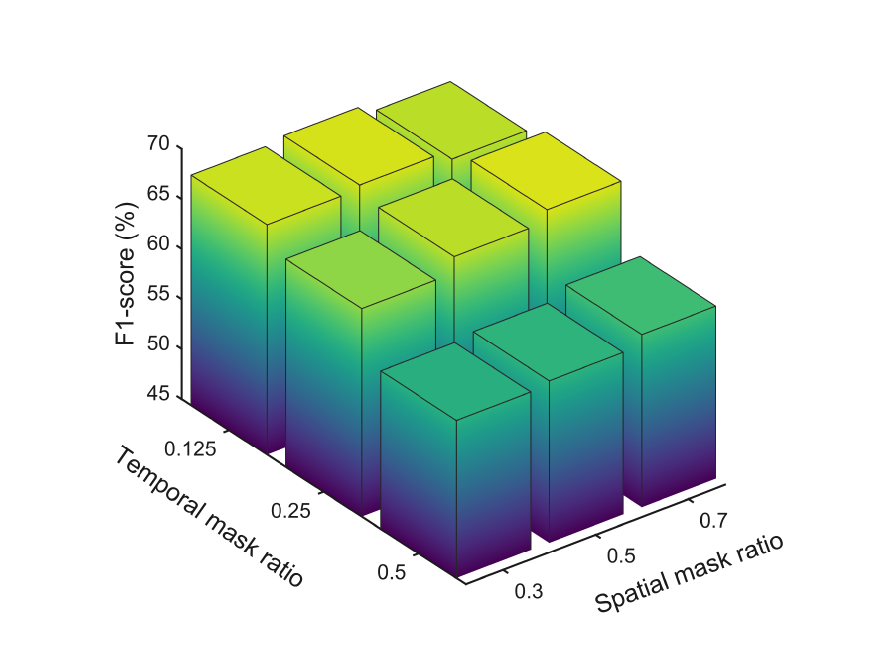} \label{fig_masking_ratio_ssc}}

\end{tabular}
\caption{Effect of temporal and spatial masking ratios.}
\label{fig_masking_ratio}
\end{figure}

In this section, we systematically evaluate the impact of different temporal and spatial masking ratio combinations on model performance. Figure \ref{fig_masking_ratio} presents the average macro F1 scores across various masking ratios. Results reveal a complex interplay between the masking ratios. When the temporal masking ratio remains fixed, model performance often exhibits an inverted U-shape or a gradual increase with a rising graph masking ratio. Conversely, fixing the graph masking ratio typically leads to a performance decline as the temporal masking ratio increases. This behavior can be attributed to two competing factors. High masking ratios can impede information recovery, hindering the model's adaptability. Conversely, overly low masking ratios provide insufficient challenge, limiting the model's capacity to extract valuable information. To balance these effects and ensure practical applicability, we adopt a temporal masking ratio of $0.125$ and a graph masking ratio of $0.5$ across all datasets in our study.

\section{Conclusion}
In this paper, we introduce TERSE, a novel method for MTS-SFDA. TERSE addresses the challenge of maintaining spatial-temporal consistency in MTS across domains without access to source data. This is achieved through a customized spatial-temporal feature encoder, coupled with both temporal restoration and spatial rewiring auxiliary tasks designed to reconstruct the latent representations and spatially correlated structures of the original sequences. TERSE is the first method tailored to account for spatial-temporal dependencies of MTS data within source-free contexts. We demonstrate the efficacy and versatility of TERSE through comprehensive experiments on three real-world datasets, where it achieves significant improvements over existing methods.

\begin{acks}
This work was supported by the National Natural Science Foundation of China (Nos. 62136004, 62276130), by the National Key R\&D Program of China (No. 2023YFF1204803), and also by the National Research Foundation, Singapore under its AI Singapore Programme (AISG2-RP-2021-027).
\end{acks}

\bibliographystyle{ACM-Reference-Format}
\bibliography{main}

\appendix

\section{Descriptions of Datasets}
\label{appendix_datasets}
\subsection{UCIHAR Dataset} 
This dataset caters specifically to the task of human activity recognition. It comprises sensor data collected from 30 distinct users, each representing a separate domain for evaluation purposes. Three sensor types are employed: accelerometers, gyroscopes, and body sensors. Each sensor captures data on three axes, resulting in a total of nine channels per sample, with each channel containing 128 data points. To assess the model's cross-domain performance, five experiments are conducted. In each experiment, the model is trained on data from one user and then tested on data collected from different users \cite{uciHAR_dataset}.

\subsection{Sleep Stage Classification (SSC) Dataset} 
The Sleep-EDF dataset provides multi-channel time series data for sleep stage classification. The data originates from six channels recording various physiological signals: including EEG (Epz-Cz, Pz-Oz), EOG, and EMG. The dataset comprises recordings from 20 subjects, each segmented and classified into five sleep stages: wake, light sleep stage 1 (N1), light sleep stage 2 (N2), deep sleep stage 3 (N3), and rapid eye movement (REM) \cite{sleepEDF_dataset}. To facilitate the evaluation of cross-domain generalizability, we adopt the same cross-scenarios as in the previous study \cite{MAPU}, treating each subject as a separate domain and selecting five cross-domain scenarios. Notably, each sample in the dataset represents a 30-second window of physiological data captured at a sampling rate of 100 Hz, resulting in a total of 3000 timestamps per sample.

\subsection{Wireless Sensor Data Mining (WISDM) Dataset}
This dataset complements the UCIHAR dataset by providing human activity recognition data collected using accelerometer sensors from 36 subjects. The subjects performed the same set of activities as those in the UCIHAR dataset. Each sample contains data from three channels (corresponding to the three accelerometer axes), with each channel having 128 data points. However, this dataset presents a potentially greater challenge due to class imbalance issues within individual subject data. We treat each subject's data as a separate domain to evaluate cross-domain generalizability. Five cross-domain scenarios are formulated by randomly selecting subjects \cite{wisdm_dataset}.

More details about the datasets are included in Table \ref{tbl:datasets}.

\input{Tables/datasets/datasets}

\section{Descriptions of Baseline Methods}
\label{appendix_baselines}
\subsection{Conventional UDA methods}
\begin{itemize}
    \item Deep Domain Confusion (DDC) \cite{ddc}: utilizes Maximum Mean Discrepancy (MMD) to minimize the difference between source and target feature distributions, aiming for domain-invariant representations.
    \item Domain-Adversarial Training of Neural Networks (DANN) \cite{DANN}: employs adversarial training to push the encoder network towards learning domain-agnostic representations.
    \item Deep Correlation Alignment (DCORAL) \cite{deep_coral}: focuses on aligning second-order statistics (variances and covariances) of source and target domains for effective adaptation.
    \item Conditional Domain Adversarial Network (CDAN) \cite{CDAN}: introduces conditional adversarial alignment, incorporating task-specific knowledge for improved adaptation.
     \item Minimum Discrepancy Estimation for Deep Domain Adaptation (MMDA) \cite{MMDA}: combines MMD, correlation alignment, and entropy minimization for comprehensive domain adaptation.
    \item Convolutional deep adaptation for time series (CoDATS) \cite{codats}: leverages adversarial training to enhance adaptation performance specifically on time series data with weak supervision.
    \item Adversarial spectral kernel matching (AdvSKM)\cite{dskn}: uses adversarial spectral kernel matching to cater to challenges of non-stationarity and non-monotonicity in time series data.
\end{itemize}

\subsection{Source-free DA methods}
\begin{itemize}
    \item Source Hypothesis Transfer (SHOT) \cite{shot}: maximizes mutual information loss with self-supervised pseudo-labels to identify target features compatible with the source hypothesis, enabling adaptation without source labels.
    \item Exploiting the intrinsic neighborhood structure (NRC) \cite{nrc}: captures the intrinsic structure of target data by forming clear clusters and enforcing label consistency among similar data points, addressing the lack of labeled target data.
    \item Attracting and dispersing (AaD) \cite{aad}: optimizes prediction consistency, encouraging similar predictions for neighboring features in the space. This leverages the inherent structure within the unlabeled target data.
    \item SF(DA)$^2$ \cite{sfda2}: adapts to target domains without source data by constructing an augmentation graph in the feature space and applying spectral neighborhood clustering, while using implicit feature augmentation and feature disentanglement to leverage class semantics and simulate unlimited augmented features without increased computational costs.
    \item Mask and impute (MAPU) \cite{MAPU}: applies random masking to time-series signals and recovers the originals in the embedding space using a temporal imputer. This captures temporal information from the source domain.
\end{itemize}

\section{Ablation Study}
\label{appendix_feature_encoder}
\subsection{Impact of Spatial-Temporal Feature Encoder}
To evaluate the effectiveness of the customized spatial-temporal feature encoder, we conduct another ablation study examining alternative feature extraction methods. We compare the performance of a temporal-only CNN encoder and a spatial-only GNN encoder against our proposed model on five cross-domain scenarios across three datasets, as illustrated in Table \ref{tab:ablation_encoder}. The temporal encoder achieves superior recognition performance compared to the spatial encoder. This suggests that relying solely on spatial features loses critical temporal information inherent in time series data, hindering the model's ability to learn discriminative representations. Notably, our combined spatial-temporal encoder consistently achieves the best performance. This highlights the crucial role of effectively capturing spatial-temporal features for accurate classification of MTS data.
\input{Tables/abalation/encoder_abalation}

\input{Tables/abalation/ucihar_details_masking_ablation}

\input{Tables/abalation/eeg_details_masking_ablation}

\input{Tables/abalation/wisdm_details_masking_ablation}

\input{Tables/abalation/ucihar_details_encoder_ablation}

\input{Tables/abalation/eeg_details_encoder_ablation}

\input{Tables/abalation/wisdm_details_encoder_ablation}

\input{Tables/versatility/ucihar_details_versatility}

\input{Tables/versatility/eeg_details_versatility}

\input{Tables/versatility/wisdm_details_versatility}

\section{Detailed Results of Ablation Study and Model Versatility Analysis}
\label{appendix_detailed_ablation_versatility}
In this section, we present detailed results of our ablation experiments and model versatility analyses across three datasets for five cross-domain scenarios. Tables \ref{tab_har_details_ablation_auxiliary_tasks}, \ref{tab_eeg_details_ablation_auxiliary_tasks}, and \ref{tab_wisdm_details_ablation_auxiliary_tasks} provide comprehensive results of the ablation study on temporal restoration and spatial rewiring for all three datasets, measured by macro F1-score. Our method achieves optimal results in four out of five cross-domain scenarios on the UCIHAR dataset, three out of five on the SSC dataset, and two out of five on the WISDM dataset. These findings underscore the significant role of our proposed temporal restoration and spatial rewiring auxiliary tasks in enhancing performance.

Tables \ref{tab_har_details_ablation_feature_encoder}, \ref{tab_eeg_details_ablation_feature_encoder}, and \ref{tab_wisdm_details_ablation_feature_encoder} detail the ablation study results for the feature encoder across the three datasets, again using macro F1-score. Our proposed spatio-temporal feature encoder achieves optimal performance in three out of five cross-domain scenarios on the UCIHAR dataset, four out of five on the SSC dataset, and two out of five on the WISDM dataset. These results confirm that the spatio-temporal feature encoder captures richer features, thereby enhancing the model's representation and improving feature discriminative properties.

Tables \ref{tab_har_details_versatility}, \ref{tab_eeg_details_versatility}, and \ref{tab_wisdm_details_versatility} present the results of our model versatility analysis across the three datasets, measured by macro F1-score. The results demonstrate that integrating our proposed temporal restoration and spatial rewiring tasks into various SFDA methods leads to performance improvements in most scenarios. This further validates the effectiveness and versatility of our two auxiliary tasks.

\end{document}

%% file: Tables/main/ucihar_main_results.tex
\begin{table*}[h]
\centering
\setlength{\tabcolsep}{2.5mm}{
\renewcommand\arraystretch{0.95}
\begin{NiceTabular}{@{}l|c|ccccc|c@{}} 
\toprule 
Algorithm & SF & 12$\rightarrow$16 & 2$\rightarrow$11 & 6$\rightarrow$23 & 7$\rightarrow$13 & 9$\rightarrow$18 & AVG\\ \midrule
Source & \bmark & 77.85$\pm$07.41 & 96.48$\pm$02.50 & 77.76$\pm$11.37 & 85.24$\pm$09.92 & 69.61$\pm$10.69 &   81.39 \\
Target & \bmark & 99.50$\pm$00.71 & 100.0$\pm$00.00 & 100.0$\pm$00.00 & 99.64$\pm$00.51 & 99.34$\pm$00.93 &   99.70 \\ \midrule
DCORAL \cite{deep_coral} & \xmark & 86.42$\pm$04.27 & 97.48$\pm$03.55 & 84.92$\pm$05.25 & 92.37$\pm$02.95 & 83.22$\pm$04.37 &   88.88 \\
DDC \cite{ddc} & \xmark & 87.97$\pm$02.45 & \textbf{100.0$\pm$00.00} & 90.78$\pm$03.34 & 94.50$\pm$04.06 & 84.33$\pm$02.39 &  91.52 \\
MMDA \cite{MMDA} & \xmark & \underline{88.86$\pm$07.12} & \textbf{100.0$\pm$00.00} & 97.82$\pm$01.89 & 93.06$\pm$05.32 & 82.34$\pm$03.07 &  92.42 \\
DANN \cite{DANN} & \xmark & 86.04$\pm$02.18 & \textbf{100.0$\pm$00.00} & 95.38$\pm$02.34 & 92.60$\pm$01.27 & 94.07$\pm$02.51 &  93.62 \\
CDAN \cite{CDAN} & \xmark & 87.99$\pm$00.90 & \textbf{100.0$\pm$00.00} & 95.86$\pm$00.81 & 94.85$\pm$02.63 & 93.13$\pm$03.45 &  94.37 \\
CoDATS \cite{codats} & \xmark & 84.40$\pm$01.55 & \textbf{100.0$\pm$00.00} & 97.16$\pm$02.65 & 94.49$\pm$02.00 & 90.89$\pm$05.17 &  93.39 \\
AdvSKM \cite{dskn} & \xmark & 88.23$\pm$02.66 & 97.14$\pm$03.37 & 93.25$\pm$03.18 & 92.29$\pm$00.06 & 81.34$\pm$07.92 &  90.45 \\

\midrule

SHOT \cite{shot} & \cmark & 87.10$\pm$00.84 & \textbf{100.0$\pm$00.00} & 98.37$\pm$01.33 & \underline{98.97$\pm$00.86} & 94.40$\pm$07.92 &  95.77 \\
NRC \cite{nrc} & \cmark & 86.17$\pm$04.48 & \textbf{100.0$\pm$00.00} & 94.06$\pm$01.74 & 92.97$\pm$00.51 & 90.24$\pm$03.43 &  92.69 \\
AaD \cite{aad} & \cmark & \textbf{89.01$\pm$01.22} & \textbf{100.0$\pm$00.00} & 97.82$\pm$01.54 & 97.50$\pm$02.10 & \underline{95.97$\pm$00.77} &  96.06 \\
SF(DA)$^2$ \cite{sfda2} & \cmark & {88.14$\pm$00.00} & \textbf{100.0$\pm$00.00} & 96.73$\pm$00.00 & 93.01$\pm$00.57 & {95.43$\pm$00.00} &  94.66 \\
MAPU \cite{MAPU} & \cmark & 86.76$\pm$00.98 & \textbf{100.0$\pm$00.00} & \underline{98.64$\pm$00.77} & 97.45$\pm$02.94 & \textbf{100.0$\pm$00.00} &  \underline{96.57} \\

\midrule

\textbf{TERSE (Ours)} & \cmark & 88.14$\pm$00.00 & \underline{99.63$\pm$00.52} & \textbf{99.73$\pm$00.38} & \textbf{100.0$\pm$00.00} & \textbf{100.0$\pm$00.00} &  \textbf{97.50} \\
\bottomrule
\end{NiceTabular}
}
\caption{Detailed results of the five UCIHAR cross-domain scenarios in terms of macro F1-score.}
\label{table:har}
\end{table*}

%% file: Tables/main/wisdm_main_results.tex
\begin{table*}[h]
\centering
\setlength{\tabcolsep}{2.5mm}{
\renewcommand\arraystretch{0.95}
\begin{NiceTabular}{@{}l|c|ccccc|c@{}} 
\toprule 
Algorithm & SF & 28$\rightarrow$4 & 2$\rightarrow$11 & 33$\rightarrow$12 & 5$\rightarrow$26 & 6$\rightarrow$19 & AVG\\ \midrule
Source & \bmark & 63.23$\pm$15.03 & 70.99$\pm$21.59 & 27.41$\pm$02.77 & 30.06$\pm$03.85 & 55.69$\pm$05.93 &   49.47 \\
Target & \bmark & 93.29$\pm$07.97 & 94.82$\pm$03.72 & 85.25$\pm$04.48 & 94.30$\pm$04.46 & 98.54$\pm$00.85 &   93.24 \\ \midrule
DCORAL \cite{deep_coral} & \xmark & 90.35$\pm$06.89 & \underline{74.81$\pm$03.87} & 51.44$\pm$16.16 & 29.37$\pm$05.74 & 61.57$\pm$12.57 &   61.51 \\
DDC \cite{ddc} & \xmark & 85.11$\pm$08.20 & 65.04$\pm$11.11 & 55.13$\pm$06.56 & 37.76$\pm$06.22 & 56.41$\pm$08.80 &   59.89 \\
MMDA \cite{MMDA} & \xmark & 92.02$\pm$07.73 & 73.73$\pm$01.67 & 53.69$\pm$05.60 & 29.63$\pm$01.69 & 51.98$\pm$06.83 &   60.21 \\
DANN \cite{DANN} & \xmark & \textbf{98.51$\pm$01.29} & 63.60$\pm$04.81 & 74.44$\pm$17.40 & \underline{46.43$\pm$11.74} & 55.69$\pm$01.67 & 67.73 \\
CDAN \cite{CDAN} & \xmark & 94.08$\pm$06.12 & 58.81$\pm$14.56 & 65.42$\pm$07.58 & 36.71$\pm$06.54 & 34.34$\pm$01.60 &   57.87 \\
CoDATS \cite{codats} & \xmark & 86.70$\pm$11.37 & \textbf{77.01$\pm$05.12} & 69.79$\pm$12.73 & \textbf{52.78$\pm$18.29} & 60.07$\pm$03.20 &   69.27 \\
AdvSKM \cite{dskn} & \xmark & 81.26$\pm$11.05 & 64.00$\pm$12.11 & 57.85$\pm$08.30 & 33.61$\pm$05.55 & \textbf{73.46$\pm$02.59} &   62.04 \\

\midrule

SHOT \cite{shot} & \cmark & 90.01 $\pm$04.84 & 63.12$\pm$11.01 & 68.07$\pm$09.28 & 29.73$\pm$00.35 & 61.46$\pm$00.86 &   62.48 \\
NRC \cite{nrc} & \cmark & 94.00 $\pm$01.69 & 66.62$\pm$12.22 & 55.01$\pm$10.38 & 35.13$\pm$08.22 & 63.73$\pm$08.21 &   62.90 \\
AaD \cite{aad} & \cmark & 87.34 $\pm$09.19 & 64.62$\pm$11.52 & 65.50$\pm$09.12 & 39.31$\pm$08.44 & \underline{70.66$\pm$07.71} &   65.49 \\
SF(DA)$^2$ \cite{sfda2} & \cmark & 81.99 $\pm$09.25 & 66.14$\pm$06.80 & 60.74$\pm$04.74 & 30.74$\pm$01.87 & {70.32$\pm$04.32} &   61.99 \\
MAPU \cite{MAPU} & \cmark & \underline{96.99 $\pm$02.61} & 74.02$\pm$07.88 & \underline{77.48$\pm$06.53} & 29.64$\pm$04.89 & 69.34$\pm$09.70 &   \underline{69.49} \\

\midrule
\textbf{TERSE (Ours)} & \cmark & 93.34 $\pm$00.91 & 74.40$\pm$00.32 & \textbf{89.73$\pm$12.29} & 41.16$\pm$01.60 & 70.60$\pm$10.13 &   \textbf{73.84} \\

\bottomrule
\end{NiceTabular}
}
\caption{Detailed results of the five WISDM cross-domain scenarios in terms of macro F1-score.}
\label{table:wisdm}
\end{table*}

%% file: Tables/main/eeg_main_results.tex
\begin{table*}[h]
\centering
\setlength{\tabcolsep}{2.5mm}{
\renewcommand\arraystretch{0.95}
\begin{NiceTabular}{@{}l|c|ccccc|c@{}} 
\toprule 
Algorithm & SF & 0$\rightarrow$11 & 12$\rightarrow$5 & 16$\rightarrow$1 & 7$\rightarrow$18 & 9$\rightarrow$14 & AVG\\ \midrule
Source & \bmark & 52.52$\pm$01.13 & 54.74$\pm$05.98 & 64.77$\pm$01.91 & 62.41$\pm$00.27 & 73.41$\pm$02.26 &  61.57 \\
Target & \bmark & 75.78$\pm$02.60 & 82.58$\pm$00.79 & 83.87$\pm$00.72 & 77.78$\pm$00.80 & 81.06$\pm$00.86 &  80.21 \\ \midrule
DCORAL \cite{deep_coral} & \xmark & 54.73$\pm$01.35 & 58.28$\pm$01.38 & 69.77$\pm$01.52 & 68.10$\pm$02.70 & 74.32$\pm$01.97 &   65.04 \\
DDC \cite{ddc} & \xmark & \textbf{56.27$\pm$01.56} & 59.21$\pm$03.95 & 71.53$\pm$04.47 & 67.87$\pm$04.30 & 72.09$\pm$04.00 &   65.40 \\
MMDA \cite{MMDA} & \xmark & 53.29$\pm$00.51 & 62.17$\pm$05.27 & 69.62$\pm$02.42 & 64.41$\pm$03.91 & 74.21$\pm$00.27 &   64.74 \\
DANN \cite{DANN} & \xmark & 54.48$\pm$02.46 & 69.99$\pm$07.42 & 67.04$\pm$03.80 & 72.87$\pm$03.11 & 74.31$\pm$00.71 &   67.74 \\
CDAN \cite{CDAN} & \xmark & 53.61$\pm$02.31 & 67.58$\pm$01.41 & 69.80$\pm$02.04 & 72.29$\pm$01.43 & \textbf{75.62$\pm$01.83} &   \underline{67.78} \\
CoDATS \cite{codats} & \xmark & 55.03$\pm$01.04 & 70.31$\pm$03.16 & 65.78$\pm$00.86 & \underline{73.87$\pm$00.52} & 73.72$\pm$01.18 &   67.74 \\
AdvSKM \cite{dskn} & \xmark & \underline{55.79$\pm$02.03} & 59.09$\pm$03.49 & 67.72$\pm$03.80 & 67.45$\pm$01.45 & 72.73$\pm$00.70 &   64.56 \\

\midrule

SHOT \cite{shot} & \cmark & 52.16 $\pm$02.59 & 60.29$\pm$01.67 & \underline{71.67$\pm$01.23} & 67.63$\pm$03.07 & 71.98$\pm$03.54 &   64.75 \\
NRC \cite{nrc} & \cmark & 53.41 $\pm$00.42 & 62.35$\pm$03.04 & 68.75$\pm$03.21 & 71.71$\pm$02.50 & \underline{75.50$\pm$02.45} &   66.34 \\
AaD \cite{aad} & \cmark & 48.99 $\pm$07.30 & 58.60$\pm$07.48 & 64.76$\pm$00.89 & 66.17$\pm$07.05 & 71.06$\pm$00.69 &   61.92 \\
SF(DA)$^2$ \cite{sfda2} & \cmark & 50.04 $\pm$03.24 & 57.32$\pm$00.97 & 66.46$\pm$01.53 & 68.42$\pm$01.73 & 66.61$\pm$00.75 &   61.77 \\
MAPU \cite{MAPU} & \cmark & 42.65 $\pm$02.32 & \textbf{75.77$\pm$02.98} & 64.78$\pm$03.73 & 72.25$\pm$01.16 & 72.74$\pm$01.33 &   65.64 \\

\midrule

\textbf{TERSE (Ours)} & \cmark & 48.30 $\pm$00.73 & \underline{71.83$\pm$06.38} & \textbf{73.69$\pm$00.28} & \textbf{75.29$\pm$01.34} & 72.89$\pm$03.11 &   \textbf{68.40} \\

\bottomrule
\end{NiceTabular}
}
\caption{Detailed results of the five SSC cross-domain scenarios in terms of macro F1-score.}
\label{table:ssc}
\end{table*}

%% file: Tables/abalation/masking_abalation.tex
\begin{table}[h]
\centering
\setlength{\tabcolsep}{2.0mm}{
  \begin{NiceTabular}{@{}cc|ccc@{}}
\toprule
Sapt Rew. & Temp Res. & UCIHAR & SSC & WISDM \\ \midrule
& & 95.77 & 64.75  & 62.48  \\ 
\cmark & & 96.16 & 66.59 & 71.19 \\
& \cmark & 96.57 & 65.64 & 69.49 \\ \midrule
\cmark & \cmark & \textbf{97.50} &\textbf{68.40} & \textbf{73.84} \\
\bottomrule
\end{NiceTabular}}

\caption{The effectiveness of temporal restoration and spatial rewiring auxiliary task. Detailed results are given in Appendix \ref{appendix_detailed_ablation_versatility}.}
\label{tab:ablation_tasks}
\end{table}

%% file: Tables/versatility/versatility_analysis.tex
\begin{table}[h]
\centering
\setlength{\tabcolsep}{3.5mm}{
\begin{NiceTabular}{c|ccc}
\toprule
Variants & UCIHAR & SSC & WISDM \\ \midrule
SHOT & 95.77 & 64.75  & 62.48  \\ 
\textbf{SHOT+} & \textbf{96.89} & \textbf{65.99} & \textbf{66.34} \\

\midrule
NRC              & 92.69 & \textbf{66.34} & 62.90 \\
\textbf{NRC+}       & \textbf{92.70} & 66.18 & \textbf{69.78} \\\midrule
AaD              & 96.06 & 61.92 & 65.49 \\
\textbf{AaD+}       & \textbf{96.64} &\textbf{63.91} & \textbf{69.81} \\
\bottomrule
\end{NiceTabular}
}

\caption{Intergrating proposed temporal restoration and spatial rewiring with existing SFDA methods on three datasets. Detailed results are given in Appendix \ref{appendix_detailed_ablation_versatility}.}
\label{tab:versatility}
\end{table}

%% file: Tables/datasets/datasets.tex
\begin{table}[!tbh]
\centering
\setlength{\tabcolsep}{1.5mm}{
\renewcommand\arraystretch{1.1}
\begin{NiceTabular}{l|ccc|cc}
\toprule
\textbf{Dataset} & \textbf{C}  & \textbf{K} & \textbf{L} & \# train samples & \# test samples \\ \midrule
UCIHAR &  9 & 6 & 128 & 2300 & 990 \\ 
SSC & 6 & 5 & 3000 & 14280 & 6130 \\ 
WISDM & 3 & 6 & 128 & 1350 & 720 \\
\bottomrule
\end{NiceTabular}
}
\caption{Details of the adopted datasets (C: \#channels, K: \#classes, L: sample length).}
\label{tbl:datasets}
\end{table}

%% file: Tables/abalation/encoder_abalation.tex
\begin{table}[h]
\centering
{\fontsize{9}{11}\selectfont
\setlength{\tabcolsep}{3.5mm}{
\renewcommand\arraystretch{1.1}
\begin{NiceTabular}{l|ccc}
\toprule
Variants & UCIHAR & SSC & WISDM \\ \midrule
Spatial GNN &  72.61 & 45.33  & 40.85  \\ 
Temporal CNN              & 89.32 & 64.39 & 60.41 \\ \midrule
\textbf{TERSE (Ours)}       & \textbf{97.50} &\textbf{68.40} & \textbf{73.84} \\
\bottomrule
\end{NiceTabular}
}}

\caption{The efficacy of temporal-spatial feature encoder. Detailed results are given in Appendix \ref{appendix_detailed_ablation_versatility}.}
\label{tab:ablation_encoder}
\end{table}

%% file: Tables/abalation/ucihar_details_masking_ablation.tex
\begin{table*}[h]
\centering
{\fontsize{9}{11}\selectfont
\setlength{\tabcolsep}{2.5mm}{
\renewcommand\arraystretch{1.1}
\begin{NiceTabular}{@{}ll|ccccc|c@{}} 
\toprule 
Sapt Rew. & Temp Res. & 12$\rightarrow$16 & 2$\rightarrow$11 & 6$\rightarrow$23 & 7$\rightarrow$13 & 9$\rightarrow$18 & AVG\\ \midrule
 &  & \underline{87.10$\pm$0.84} & \textbf{100.0$\pm$0.00} & 98.37$\pm$1.33 & \underline{98.97$\pm$0.86} & 94.40$\pm$7.92 &  95.77 \\
\cmark &  & \textbf{88.14$\pm$0.00} & \textbf{100.0$\pm$0.00} & 95.40$\pm$1.53 & 97.85$\pm$2.92 & \underline{99.42$\pm$0.51} & 96.16 \\
 & \cmark & 86.76$\pm$0.98 & \textbf{100.0$\pm$0.00} & \underline{98.64$\pm$0.77} & 97.45$\pm$2.94 & \textbf{100.0$\pm$0.00} &  \underline{96.57} \\
 \midrule
\cmark & \cmark & \textbf{88.14$\pm$0.00} & \underline{99.63$\pm$0.52} & \textbf{99.73$\pm$0.38} & \textbf{100.0$\pm$0.00} & \textbf{100.0$\pm$0.00} &  \textbf{97.50} \\

\bottomrule
\end{NiceTabular}
}}
\caption{Detailed results of ablation study of temporal restoration and spatial rewiring on the five UCIHAR cross-domain scenarios in terms of macro F1-score.}
\label{tab_har_details_ablation_auxiliary_tasks}
\end{table*}

%% file: Tables/abalation/eeg_details_masking_ablation.tex
\begin{table*}[h]
\centering
{\fontsize{9}{11}\selectfont
\setlength{\tabcolsep}{2.5mm}{
\renewcommand\arraystretch{1.1}
\begin{NiceTabular}{@{}ll|ccccc|c@{}} 
\toprule 
Sapt Rew. & Temp Res. & 0$\rightarrow$11 & 12$\rightarrow$5 & 16$\rightarrow$1 & 7$\rightarrow$18 & 9$\rightarrow$14 & AVG\\ \midrule
 &  & \textbf{52.16 $\pm$2.59} & 60.29$\pm$1.67 & \underline{71.67$\pm$1.23} & 67.63$\pm$3.07 & 71.98$\pm$3.54 &   64.75 \\
\cmark &  & \underline{50.06$\pm$0.49} & \underline{73.10$\pm$2.08} & 68.65$\pm$0.82 & 71.09$\pm$0.51 & 70.07$\pm$1.82 & \underline{66.59} \\
 & \cmark & 42.65 $\pm$2.32 & \textbf{75.77$\pm$2.98} & 64.78$\pm$3.73 & \underline{72.25$\pm$1.16} & \underline{72.74$\pm$1.33} &   65.64 \\
 \midrule
\cmark & \cmark & 48.30 $\pm$0.73 & 71.83$\pm$6.38 & \textbf{73.69$\pm$0.28} & \textbf{75.29$\pm$1.34} & \textbf{72.89$\pm$3.11} &   \textbf{68.40} \\

\bottomrule
\end{NiceTabular}
}}
\caption{Detailed results of ablation study of temporal restoration and spatial rewiring on the five SSC cross-domain scenarios in terms of macro F1-score.}
\label{tab_eeg_details_ablation_auxiliary_tasks}
\end{table*}

%% file: Tables/abalation/wisdm_details_masking_ablation.tex
\begin{table*}[h]
\centering
{\fontsize{9}{11}\selectfont
\setlength{\tabcolsep}{2.5mm}{
\renewcommand\arraystretch{1.1}
\begin{NiceTabular}{@{}ll|ccccc|c@{}} 
\toprule 
Sapt Rew. & Temp Res. & 28$\rightarrow$4 & 2$\rightarrow$11 & 33$\rightarrow$12 & 5$\rightarrow$26 & 6$\rightarrow$19 & AVG\\ \midrule
 &  & 90.01 $\pm$4.84 & 63.12$\pm$11.01 & 68.07$\pm$09.28 & 29.73$\pm$0.35 & 61.46$\pm$00.86 &   62.48 \\
\cmark &  & \textbf{98.11$\pm$2.58} & \textbf{79.75$\pm$06.41} & 63.57$\pm$12.62 & \underline{36.54$\pm$4.07} & \textbf{77.96$\pm$01.63} & \underline{71.19} \\
 & \cmark & \underline{96.99 $\pm$2.61} & 74.02$\pm$07.88 & \underline{77.48$\pm$06.53} & 29.64$\pm$4.89 & 69.34$\pm$09.70 &   69.49 \\
 \midrule
\cmark & \cmark & 93.34 $\pm$0.91 & \underline{74.40$\pm$00.32} & \textbf{89.73$\pm$12.29} & \textbf{41.16$\pm$1.60} & \underline{70.60$\pm$10.13} &   \textbf{73.84} \\

\bottomrule
\end{NiceTabular}
}}
\caption{Detailed results of ablation study of temporal restoration and spatial rewiring on the five WISDM cross-domain scenarios in terms of macro F1-score.}
\label{tab_wisdm_details_ablation_auxiliary_tasks}
\end{table*}

%% file: Tables/abalation/ucihar_details_encoder_ablation.tex
\begin{table*}[h]
\centering
{\fontsize{9}{11}\selectfont
\setlength{\tabcolsep}{2.5mm}{
\renewcommand\arraystretch{1.1}
\begin{NiceTabular}{@{}l|ccccc|c@{}} 
\toprule 
Variants & 12$\rightarrow$16 & 2$\rightarrow$11 & 6$\rightarrow$23 & 7$\rightarrow$13 & 9$\rightarrow$18 & AVG\\ \midrule
Spatial GNN &  \underline{69.62$\pm$8.80} & 82.08$\pm$1.12 & 63.04$\pm$1.99 & 85.83$\pm$3.68 & 62.48$\pm$3.58 & 72.61 \\
Temporal CNN &  68.01$\pm$2.27 & \textbf{100.0$\pm$0.00} & \textbf{100.0$\pm$0.00} & \underline{97.78 $\pm$3.85} & \underline{80.83$\pm$5.07} & \underline{89.32} \\
 \midrule
\textbf{TERSE (Ours)} & \textbf{88.14$\pm$0.00} & \underline{99.63$\pm$0.52} & \underline{99.73$\pm$0.38} & \textbf{100.0$\pm$0.00} & \textbf{100.0$\pm$0.00} &  \textbf{97.50} \\

\bottomrule
\end{NiceTabular}
}}
\caption{Detailed results of ablation study of feature encoder on the five UCIHAR cross-domain scenarios in terms of macro F1-score.}
\label{tab_har_details_ablation_feature_encoder}
\end{table*}

%% file: Tables/abalation/eeg_details_encoder_ablation.tex
\begin{table*}[h]
\centering
{\fontsize{9}{11}\selectfont
\setlength{\tabcolsep}{2.5mm}{
\renewcommand\arraystretch{1.1}
\begin{NiceTabular}{@{}l|ccccc|c@{}} 
\toprule 
Variants & 0$\rightarrow$11 & 12$\rightarrow$5 & 16$\rightarrow$1 & 7$\rightarrow$18 & 9$\rightarrow$14 & AVG\\ \midrule
 Spatial GNN & 42.99$\pm$00.96 & 38.45$\pm$2.46 & 44.73$\pm$2.01 & 48.61$\pm$3.46 & 51.89$\pm$1.76 & 45.33 \\
Temporal CNN  & \textbf{49.00$\pm$11.96} & \underline{67.89$\pm$1.33} & \underline{64.88$\pm$1.72} & \underline{69.91$\pm$2.44} & \underline{70.26$\pm$3.46} & \underline{64.39} \\
\midrule
\textbf{TERSE (Ours)} & \underline{48.30 $\pm$00.73} & \textbf{71.83$\pm$6.38} & \textbf{73.69$\pm$0.28} & \textbf{75.29$\pm$1.34} & \textbf{72.89$\pm$3.11} &   \textbf{68.40} \\

\bottomrule
\end{NiceTabular}
}}
\caption{Detailed results of ablation study of feature encoder on the five SSC cross-domain scenarios in terms of macro F1-score.}
\label{tab_eeg_details_ablation_feature_encoder}
\end{table*}

%% file: Tables/abalation/wisdm_details_encoder_ablation.tex
\begin{table*}[h]
\centering
{\fontsize{9}{11}\selectfont
\setlength{\tabcolsep}{2.5mm}{
\renewcommand\arraystretch{1.1}
\begin{NiceTabular}{@{}l|ccccc|c@{}} 
\toprule 
Variants & 28$\rightarrow$4 & 2$\rightarrow$11 & 33$\rightarrow$12 & 5$\rightarrow$26 & 6$\rightarrow$19 & AVG\\ \midrule
Spatial GNN & 50.85$\pm$02.69 & 39.72$\pm$02.81 & 30.77$\pm$03.20 & \textbf{41.63$\pm$03.75} & 41.28$\pm$00.21 & 40.85 \\
Temporal CNN  & 71.22$\pm$12.19 & \textbf{80.78$\pm$12.43} & \underline{60.97$\pm$09.87} & 35.20$\pm$10.98 & \underline{53.87$\pm$08.09} & \underline{60.41} \\
\midrule
\textbf{TERSE (Ours)} & \textbf{93.34 $\pm$00.91} & \underline{74.40$\pm$00.32} & \textbf{89.73$\pm$12.29} & \underline{41.16$\pm$01.60} & \underline{70.60$\pm$10.13} &   \textbf{73.84} \\

\bottomrule
\end{NiceTabular}
}}
\caption{Detailed results of ablation study of feature encoder on the five WISDM cross-domain scenarios in terms of macro F1-score.}
\label{tab_wisdm_details_ablation_feature_encoder}
\end{table*}

%% file: Tables/versatility/ucihar_details_versatility.tex
\begin{table*}[h]
\centering
{\fontsize{9}{11}\selectfont
\setlength{\tabcolsep}{2.5mm}{
\renewcommand\arraystretch{1.1}
\begin{NiceTabular}{@{}l|ccccc|c@{}} 
\toprule 
Variants & 12$\rightarrow$16 & 2$\rightarrow$11 & 6$\rightarrow$23 & 7$\rightarrow$13 & 9$\rightarrow$18 & AVG\\ \midrule
SHOT &  87.10$\pm$0.84 & \textbf{100.0$\pm$0.00} & \textbf{98.37$\pm$1.33} & 98.97$\pm$0.86 & 94.40$\pm$7.92 & 95.77 \\
\textbf{SHOT+} &  \textbf{87.63$\pm$0.89} & \textbf{100.0$\pm$0.00} & 97.82$\pm$1.89 & \textbf{99.29$\pm$1.22} & \textbf{99.71$\pm$0.51} & \textbf{96.89} \\
\midrule
NRC & 86.17$\pm$4.48 & \textbf{100.0$\pm$0.00} & 94.06$\pm$1.74 & \textbf{92.97$\pm$0.51} & 90.24$\pm$3.43 & 92.69 \\
\textbf{NRC+} & \textbf{86.65$\pm$1.72} & 98.13$\pm$3.24 & \textbf{95.73$\pm$1.74} & 92.62$\pm$1.24 & \textbf{90.38$\pm$5.28} & \textbf{92.70} \\
\midrule
AaD & \textbf{89.01$\pm$1.22} & \textbf{100.0$\pm$0.00} & \textbf{97.82$\pm$1.54} & 97.50$\pm$2.10 & 95.97$\pm$0.77 & 96.06 \\
\textbf{AaD+} & 88.14$\pm$0.00 & 98.13$\pm$3.24 & 97.23$\pm$1.76 & \textbf{100.00$\pm$0.00} & \textbf{99.71$\pm$0.51} & \textbf{96.64} \\

\bottomrule
\end{NiceTabular}
}}
\caption{Detailed results of model versatility on the five UCIHAR cross-domain scenarios in terms of macro F1-score.}
\label{tab_har_details_versatility}
\end{table*}

%% file: Tables/versatility/eeg_details_versatility.tex
\begin{table*}[h]
\centering
{\fontsize{9}{11}\selectfont
\setlength{\tabcolsep}{2.5mm}{
\renewcommand\arraystretch{1.1}
\begin{NiceTabular}{@{}l|ccccc|c@{}} 
\toprule 
Variants & 0$\rightarrow$11 & 12$\rightarrow$5 & 16$\rightarrow$1 & 7$\rightarrow$18 & 9$\rightarrow$14 & AVG\\ \midrule
SHOT &  \textbf{52.16$\pm$2.59} & 60.29$\pm$1.67 & \textbf{71.67$\pm$1.23} & 67.63$\pm$3.07 & \textbf{71.98$\pm$3.54} & 64.75 \\
\textbf{SHOT+} &  50.30$\pm$1.63 & \textbf{65.72$\pm$4.39} & 68.11$\pm$2.50 & \textbf{74.69$\pm$0.66} & 71.14$\pm$1.99 & \textbf{65.99} \\
\midrule
NRC & \textbf{53.41$\pm$0.42} & 62.35$\pm$3.04 & \textbf{68.75$\pm$3.21} & 71.71$\pm$2.50 & \textbf{75.50$\pm$2.45} & \textbf{66.34} \\
\textbf{NRC+} & 49.73$\pm$2.90 & \textbf{72.11$\pm$2.55} & 66.38$\pm$3.39 & \textbf{73.39$\pm$2.38} & 69.31$\pm$2.96 & 66.18 \\
\midrule
AaD & \textbf{48.99$\pm$7.30} & 58.60$\pm$7.48 & 64.76$\pm$0.89 & 66.17$\pm$7.05 & \textbf{71.06$\pm$0.69} & 61.92 \\
\textbf{AaD+} & 47.10$\pm$2.60 & \textbf{67.16$\pm$6.11} & \textbf{67.97$\pm$3.77} & \textbf{73.08$\pm$0.56} & 64.23$\pm$1.13 & \textbf{63.91} \\

\bottomrule
\end{NiceTabular}
}}
\caption{Detailed results of model versatility on the five SSC cross-domain scenarios in terms of macro F1-score.}
\label{tab_eeg_details_versatility}
\end{table*}

%% file: Tables/versatility/wisdm_details_versatility.tex
\begin{table*}[h]
\centering
{\fontsize{9}{11}\selectfont
\setlength{\tabcolsep}{2.5mm}{
\renewcommand\arraystretch{1.1}
\begin{NiceTabular}{@{}l|ccccc|c@{}} 
\toprule 
Variants & 28$\rightarrow$4 & 2$\rightarrow$11 & 33$\rightarrow$12 & 5$\rightarrow$26 & 6$\rightarrow$19 & AVG\\ \midrule
SHOT &  \textbf{90.01$\pm$04.84} & 63.12$\pm$11.01 & 68.07$\pm$09.28 & 29.73$\pm$0.35 & \textbf{61.46$\pm$0.86} & 62.48 \\
\textbf{SHOT+} &  87.85$\pm$07.90 & \textbf{73.30$\pm$04.32} & \textbf{74.40$\pm$02.35} & \textbf{36.41$\pm$5.37} & 59.72$\pm$4.22 & \textbf{66.34} \\
\midrule
NRC & \textbf{94.00$\pm$01.69} & 66.62$\pm$12.22 & 55.01$\pm$10.38 & \textbf{35.13$\pm$8.22} & 63.73$\pm$8.21 & 62.90 \\
\textbf{NRC+} & 92.03$\pm$11.90 & \textbf{79.49$\pm$10.39} & \textbf{73.75$\pm$14.36} & 33.70$\pm$7.88 & \textbf{69.91$\pm$2.97} & \textbf{69.78} \\
\midrule
AaD & 87.34$\pm$09.19 & 64.62$\pm$11.52 & 65.50$\pm$09.12 & \textbf{39.31$\pm$8.44} & \textbf{70.66$\pm$7.71} & 65.49 \\
\textbf{AaD+} & \textbf{91.78$\pm$05.20} & \textbf{76.89$\pm$08.51} & \textbf{79.40$\pm$15.10} & 39.15$\pm$7.83 & 61.81$\pm$1.65 & \textbf{69.81} \\

\bottomrule
\end{NiceTabular}
}}
\caption{Detailed results of ablation study of feature encoder on the five WISDM cross-domain scenarios in terms of macro F1-score.}
\label{tab_wisdm_details_versatility}
\end{table*}

%% file: main.bbl

\begin{thebibliography}{61}


\ifx \showCODEN    \undefined \def \showCODEN     #1{\unskip}     \fi
\ifx \showDOI      \undefined \def \showDOI       #1{#1}\fi
\ifx \showISBNx    \undefined \def \showISBNx     #1{\unskip}     \fi
\ifx \showISBNxiii \undefined \def \showISBNxiii  #1{\unskip}     \fi
\ifx \showISSN     \undefined \def \showISSN      #1{\unskip}     \fi
\ifx \showLCCN     \undefined \def \showLCCN      #1{\unskip}     \fi
\ifx \shownote     \undefined \def \shownote      #1{#1}          \fi
\ifx \showarticletitle \undefined \def \showarticletitle #1{#1}   \fi
\ifx \showURL      \undefined \def \showURL       {\relax}        \fi
\providecommand\bibfield[2]{#2}
\providecommand\bibinfo[2]{#2}
\providecommand\natexlab[1]{#1}
\providecommand\showeprint[2][]{arXiv:#2}

\bibitem[Ahmed et~al\mbox{.}(2021)]%
        {Entropymin2}
\bibfield{author}{\bibinfo{person}{Sk~Miraj Ahmed}, \bibinfo{person}{Dripta~S Raychaudhuri}, \bibinfo{person}{Sujoy Paul}, \bibinfo{person}{Samet Oymak}, {and} \bibinfo{person}{Amit~K Roy-Chowdhury}.} \bibinfo{year}{2021}\natexlab{}.
\newblock \showarticletitle{Unsupervised multi-source domain adaptation without access to source data}. In \bibinfo{booktitle}{\emph{Proceedings of the IEEE/CVF Conference on Computer Vision and Pattern Recognition}}. \bibinfo{pages}{10103--10112}.
\newblock


\bibitem[{Anguita} et~al\mbox{.}(2013)]%
        {uciHAR_dataset}
\bibfield{author}{\bibinfo{person}{Davide {Anguita}}, \bibinfo{person}{Alessandro {Ghio}}, \bibinfo{person}{Luca {Oneto}}, \bibinfo{person}{Xavier {Parra}}, {and} \bibinfo{person}{Jorge~Luis {Reyes-Ortiz}}.} \bibinfo{year}{2013}\natexlab{}.
\newblock \showarticletitle{A public domain dataset for human activity recognition using smartphones}. In \bibinfo{booktitle}{\emph{European Symposium on Artificial Neural Networks}}.
\newblock


\bibitem[Cai et~al\mbox{.}(2021)]%
        {SASA}
\bibfield{author}{\bibinfo{person}{Ruichu Cai}, \bibinfo{person}{Jiawei Chen}, \bibinfo{person}{Zijian Li}, \bibinfo{person}{Wei Chen}, \bibinfo{person}{Keli Zhang}, \bibinfo{person}{Junjian Ye}, \bibinfo{person}{Zhuozhang Li}, \bibinfo{person}{Xiaoyan Yang}, {and} \bibinfo{person}{Zhenjie Zhang}.} \bibinfo{year}{2021}\natexlab{}.
\newblock \showarticletitle{Time series domain adaptation via sparse associative structure alignment}. In \bibinfo{booktitle}{\emph{Proceedings of the AAAI Conference on Artificial Intelligence}}, Vol.~\bibinfo{volume}{35}. \bibinfo{pages}{6859--6867}.
\newblock


\bibitem[Chu et~al\mbox{.}(2022)]%
        {vc3}
\bibfield{author}{\bibinfo{person}{Tong Chu}, \bibinfo{person}{Yahao Liu}, \bibinfo{person}{Jinhong Deng}, \bibinfo{person}{Wen Li}, {and} \bibinfo{person}{Lixin Duan}.} \bibinfo{year}{2022}\natexlab{}.
\newblock \showarticletitle{Denoised maximum classifier discrepancy for source-free unsupervised domain adaptation}. In \bibinfo{booktitle}{\emph{Proceedings of the AAAI Conference on Artificial Intelligence}}, Vol.~\bibinfo{volume}{36}. \bibinfo{pages}{472--480}.
\newblock


\bibitem[Ding et~al\mbox{.}(2022)]%
        {Select3}
\bibfield{author}{\bibinfo{person}{Ning Ding}, \bibinfo{person}{Yixing Xu}, \bibinfo{person}{Yehui Tang}, \bibinfo{person}{Chao Xu}, \bibinfo{person}{Yunhe Wang}, {and} \bibinfo{person}{Dacheng Tao}.} \bibinfo{year}{2022}\natexlab{}.
\newblock \showarticletitle{Source-free domain adaptation via distribution estimation}. In \bibinfo{booktitle}{\emph{Proceedings of the IEEE/CVF Conference on Computer Vision and Pattern Recognition}}. \bibinfo{pages}{7212--7222}.
\newblock


\bibitem[Ding et~al\mbox{.}(2023)]%
        {Pseudolabel2}
\bibfield{author}{\bibinfo{person}{Yuhe Ding}, \bibinfo{person}{Lijun Sheng}, \bibinfo{person}{Jian Liang}, \bibinfo{person}{Aihua Zheng}, {and} \bibinfo{person}{Ran He}.} \bibinfo{year}{2023}\natexlab{}.
\newblock \showarticletitle{Proxymix: Proxy-based mixup training with label refinery for source-free domain adaptation}.
\newblock \bibinfo{journal}{\emph{Neural Networks}}  \bibinfo{volume}{167} (\bibinfo{year}{2023}), \bibinfo{pages}{92--103}.
\newblock


\bibitem[Du et~al\mbox{.}(2024)]%
        {Select1}
\bibfield{author}{\bibinfo{person}{Yuntao Du}, \bibinfo{person}{Haiyang Yang}, \bibinfo{person}{Mingcai Chen}, \bibinfo{person}{Hongtao Luo}, \bibinfo{person}{Juan Jiang}, \bibinfo{person}{Yi Xin}, {and} \bibinfo{person}{Chongjun Wang}.} \bibinfo{year}{2024}\natexlab{}.
\newblock \showarticletitle{Generation, augmentation, and alignment: A pseudo-source domain based method for source-free domain adaptation}.
\newblock \bibinfo{journal}{\emph{Machine Learning}} \bibinfo{volume}{113}, \bibinfo{number}{6} (\bibinfo{year}{2024}), \bibinfo{pages}{3611--3631}.
\newblock


\bibitem[Eldele et~al\mbox{.}(2023)]%
        {CoTMix}
\bibfield{author}{\bibinfo{person}{Emadeldeen Eldele}, \bibinfo{person}{Mohamed Ragab}, \bibinfo{person}{Zhenghua Chen}, \bibinfo{person}{Min Wu}, \bibinfo{person}{Chee-Keong Kwoh}, {and} \bibinfo{person}{Xiaoli Li}.} \bibinfo{year}{2023}\natexlab{}.
\newblock \showarticletitle{Contrastive domain adaptation for time-series via temporal mixup}.
\newblock \bibinfo{journal}{\emph{IEEE Transactions on Artificial Intelligence}} (\bibinfo{year}{2023}).
\newblock


\bibitem[Eldele et~al\mbox{.}(2022)]%
        {ADAST}
\bibfield{author}{\bibinfo{person}{Emadeldeen Eldele}, \bibinfo{person}{Mohamed Ragab}, \bibinfo{person}{Zhenghua Chen}, \bibinfo{person}{Min Wu}, \bibinfo{person}{Chee-Keong Kwoh}, \bibinfo{person}{Xiaoli Li}, {and} \bibinfo{person}{Cuntai Guan}.} \bibinfo{year}{2022}\natexlab{}.
\newblock \showarticletitle{ADAST: Attentive cross-domain EEG-based sleep staging framework with iterative self-training}.
\newblock \bibinfo{journal}{\emph{IEEE Transactions on Emerging Topics in Computational Intelligence}} \bibinfo{volume}{7}, \bibinfo{number}{1} (\bibinfo{year}{2022}), \bibinfo{pages}{210--221}.
\newblock


\bibitem[Ganin et~al\mbox{.}(2016)]%
        {DANN}
\bibfield{author}{\bibinfo{person}{Yaroslav Ganin}, \bibinfo{person}{Evgeniya Ustinova}, \bibinfo{person}{Hana Ajakan}, \bibinfo{person}{Pascal Germain}, \bibinfo{person}{Hugo Larochelle}, \bibinfo{person}{Fran{\c{c}}ois Laviolette}, \bibinfo{person}{Mario March}, {and} \bibinfo{person}{Victor Lempitsky}.} \bibinfo{year}{2016}\natexlab{}.
\newblock \showarticletitle{Domain-adversarial training of neural networks}.
\newblock \bibinfo{journal}{\emph{Journal of Machine Learning Research}} \bibinfo{volume}{17}, \bibinfo{number}{59} (\bibinfo{year}{2016}), \bibinfo{pages}{1--35}.
\newblock


\bibitem[Goldberger et~al\mbox{.}(2000)]%
        {sleepEDF_dataset}
\bibfield{author}{\bibinfo{person}{Ary~L Goldberger}, \bibinfo{person}{Luis~AN Amaral}, \bibinfo{person}{Leon Glass}, \bibinfo{person}{Jeffrey~M Hausdorff}, \bibinfo{person}{Plamen~Ch Ivanov}, \bibinfo{person}{Roger~G Mark}, \bibinfo{person}{Joseph~E Mietus}, \bibinfo{person}{George~B Moody}, \bibinfo{person}{Chung-Kang Peng}, {and} \bibinfo{person}{H~Eugene Stanley}.} \bibinfo{year}{2000}\natexlab{}.
\newblock \showarticletitle{PhysioBank, PhysioToolkit, and PhysioNet components of a new research resource for complex physiologic signals}.
\newblock \bibinfo{journal}{\emph{Circulation}} (\bibinfo{year}{2000}).
\newblock


\bibitem[Gong et~al\mbox{.}(2023a)]%
        {gong2023astdf}
\bibfield{author}{\bibinfo{person}{Peiliang Gong}, \bibinfo{person}{Ziyu Jia}, \bibinfo{person}{Pengpai Wang}, \bibinfo{person}{Yueying Zhou}, {and} \bibinfo{person}{Daoqiang Zhang}.} \bibinfo{year}{2023}\natexlab{a}.
\newblock \showarticletitle{ASTDF-Net: Attention-based spatial-temporal dual-stream fusion network for EEG-based emotion recognition}. In \bibinfo{booktitle}{\emph{Proceedings of the 31st ACM International Conference on Multimedia}}. \bibinfo{pages}{883--892}.
\newblock


\bibitem[Gong et~al\mbox{.}(2024)]%
        {gong2024tfac}
\bibfield{author}{\bibinfo{person}{Peiliang Gong}, \bibinfo{person}{Pengpai Wang}, \bibinfo{person}{Yueying Zhou}, \bibinfo{person}{Xuyun Wen}, {and} \bibinfo{person}{Daoqiang Zhang}.} \bibinfo{year}{2024}\natexlab{}.
\newblock \showarticletitle{TFAC-Net: A temporal-frequential attentional convolutional network for driver drowsiness recognition with single-channel EEG}.
\newblock \bibinfo{journal}{\emph{IEEE Transactions on Intelligent Transportation Systems}} (\bibinfo{year}{2024}).
\newblock


\bibitem[Gong et~al\mbox{.}(2023b)]%
        {gong2023spiking}
\bibfield{author}{\bibinfo{person}{Peiliang Gong}, \bibinfo{person}{Pengpai Wang}, \bibinfo{person}{Yueying Zhou}, {and} \bibinfo{person}{Daoqiang Zhang}.} \bibinfo{year}{2023}\natexlab{b}.
\newblock \showarticletitle{A spiking neural network with adaptive graph convolution and LSTM for EEG-based brain-computer interfaces}.
\newblock \bibinfo{journal}{\emph{IEEE Transactions on Neural Systems and Rehabilitation Engineering}}  \bibinfo{volume}{31} (\bibinfo{year}{2023}), \bibinfo{pages}{1440--1450}.
\newblock


\bibitem[He et~al\mbox{.}(2023)]%
        {raincoat}
\bibfield{author}{\bibinfo{person}{Huan He}, \bibinfo{person}{Owen Queen}, \bibinfo{person}{Teddy Koker}, \bibinfo{person}{Consuelo Cuevas}, \bibinfo{person}{Theodoros Tsiligkaridis}, {and} \bibinfo{person}{Marinka Zitnik}.} \bibinfo{year}{2023}\natexlab{}.
\newblock \showarticletitle{Domain adaptation for time series under feature and label shifts}. In \bibinfo{booktitle}{\emph{International Conference on Machine Learning}}. PMLR, \bibinfo{pages}{12746--12774}.
\newblock


\bibitem[He et~al\mbox{.}(2015)]%
        {PReLU}
\bibfield{author}{\bibinfo{person}{Kaiming He}, \bibinfo{person}{Xiangyu Zhang}, \bibinfo{person}{Shaoqing Ren}, {and} \bibinfo{person}{Jian Sun}.} \bibinfo{year}{2015}\natexlab{}.
\newblock \showarticletitle{Delving deep into rectifiers: Surpassing human-level performance on imagenet classification}. In \bibinfo{booktitle}{\emph{Proceedings of the IEEE International Conference on Computer Vision}}. \bibinfo{pages}{1026--1034}.
\newblock


\bibitem[Huang et~al\mbox{.}(2021)]%
        {Contrast1}
\bibfield{author}{\bibinfo{person}{Jiaxing Huang}, \bibinfo{person}{Dayan Guan}, \bibinfo{person}{Aoran Xiao}, {and} \bibinfo{person}{Shijian Lu}.} \bibinfo{year}{2021}\natexlab{}.
\newblock \showarticletitle{Model adaptation: Historical contrastive learning for unsupervised domain adaptation without source data}.
\newblock \bibinfo{journal}{\emph{Advances in Neural Information Processing Systems}}  \bibinfo{volume}{34} (\bibinfo{year}{2021}), \bibinfo{pages}{3635--3649}.
\newblock


\bibitem[Huang et~al\mbox{.}(2022)]%
        {huang2022relative}
\bibfield{author}{\bibinfo{person}{Yi Huang}, \bibinfo{person}{Xiaoshan Yang}, \bibinfo{person}{Ji Zhang}, {and} \bibinfo{person}{Changsheng Xu}.} \bibinfo{year}{2022}\natexlab{}.
\newblock \showarticletitle{Relative alignment network for source-free multimodal video domain adaptation}. In \bibinfo{booktitle}{\emph{Proceedings of the 30th ACM International Conference on Multimedia}}. \bibinfo{pages}{1652--1660}.
\newblock


\bibitem[Hwang et~al\mbox{.}(2024)]%
        {sfda2}
\bibfield{author}{\bibinfo{person}{Uiwon Hwang}, \bibinfo{person}{Jonghyun Lee}, \bibinfo{person}{Juhyeon Shin}, {and} \bibinfo{person}{Sungroh Yoon}.} \bibinfo{year}{2024}\natexlab{}.
\newblock \showarticletitle{{SF}({DA})\${\textasciicircum}2\$: Source-free domain adaptation through the lens of data augmentation}. In \bibinfo{booktitle}{\emph{International Conference on Learning Representations}}.
\newblock


\bibitem[Jin et~al\mbox{.}(2022)]%
        {DAF}
\bibfield{author}{\bibinfo{person}{Xiaoyong Jin}, \bibinfo{person}{Youngsuk Park}, \bibinfo{person}{Danielle Maddix}, \bibinfo{person}{Hao Wang}, {and} \bibinfo{person}{Yuyang Wang}.} \bibinfo{year}{2022}\natexlab{}.
\newblock \showarticletitle{Domain adaptation for time series forecasting via attention sharing}. In \bibinfo{booktitle}{\emph{International Conference on Machine Learning}}. PMLR, \bibinfo{pages}{10280--10297}.
\newblock


\bibitem[Kipf and Welling(2017)]%
        {GNN}
\bibfield{author}{\bibinfo{person}{Thomas~N. Kipf} {and} \bibinfo{person}{Max Welling}.} \bibinfo{year}{2017}\natexlab{}.
\newblock \showarticletitle{Semi-supervised classification with graph convolutional networks}. In \bibinfo{booktitle}{\emph{International Conference on Learning Representations}}.
\newblock


\bibitem[Kouw and Loog(2019)]%
        {UDA_Survey}
\bibfield{author}{\bibinfo{person}{Wouter~M Kouw} {and} \bibinfo{person}{Marco Loog}.} \bibinfo{year}{2019}\natexlab{}.
\newblock \showarticletitle{A review of domain adaptation without target labels}.
\newblock \bibinfo{journal}{\emph{IEEE Transactions on Pattern Analysis and Machine Intelligence}} \bibinfo{volume}{43}, \bibinfo{number}{3} (\bibinfo{year}{2019}), \bibinfo{pages}{766--785}.
\newblock


\bibitem[Kundu et~al\mbox{.}(2022)]%
        {vc1}
\bibfield{author}{\bibinfo{person}{Jogendra~Nath Kundu}, \bibinfo{person}{Akshay~R Kulkarni}, \bibinfo{person}{Suvaansh Bhambri}, \bibinfo{person}{Deepesh Mehta}, \bibinfo{person}{Shreyas~Anand Kulkarni}, \bibinfo{person}{Varun Jampani}, {and} \bibinfo{person}{Venkatesh~Babu Radhakrishnan}.} \bibinfo{year}{2022}\natexlab{}.
\newblock \showarticletitle{Balancing discriminability and transferability for source-free domain adaptation}. In \bibinfo{booktitle}{\emph{International Conference on Machine Learning}}. PMLR, \bibinfo{pages}{11710--11728}.
\newblock


\bibitem[Kurmi et~al\mbox{.}(2021)]%
        {GAN2}
\bibfield{author}{\bibinfo{person}{Vinod~K Kurmi}, \bibinfo{person}{Venkatesh~K Subramanian}, {and} \bibinfo{person}{Vinay~P Namboodiri}.} \bibinfo{year}{2021}\natexlab{}.
\newblock \showarticletitle{Domain impression: A source data free domain adaptation method}. In \bibinfo{booktitle}{\emph{Proceedings of the IEEE/CVF Winter Conference on Applications of Computer Vision}}. \bibinfo{pages}{615--625}.
\newblock


\bibitem[Kwapisz et~al\mbox{.}(2011)]%
        {wisdm_dataset}
\bibfield{author}{\bibinfo{person}{Jennifer~R Kwapisz}, \bibinfo{person}{Gary~M Weiss}, {and} \bibinfo{person}{Samuel~A Moore}.} \bibinfo{year}{2011}\natexlab{}.
\newblock \showarticletitle{Activity recognition using cell phone accelerometers}.
\newblock \bibinfo{journal}{\emph{ACM SigKDD Explorations Newsletter}} \bibinfo{volume}{12}, \bibinfo{number}{2} (\bibinfo{year}{2011}), \bibinfo{pages}{74--82}.
\newblock


\bibitem[Li et~al\mbox{.}(2024)]%
        {SFDA_Survey}
\bibfield{author}{\bibinfo{person}{Jingjing Li}, \bibinfo{person}{Zhiqi Yu}, \bibinfo{person}{Zhekai Du}, \bibinfo{person}{Lei Zhu}, {and} \bibinfo{person}{Heng~Tao Shen}.} \bibinfo{year}{2024}\natexlab{}.
\newblock \showarticletitle{A comprehensive survey on source-free domain adaptation}.
\newblock \bibinfo{journal}{\emph{IEEE Transactions on Pattern Analysis and Machine Intelligence}} (\bibinfo{year}{2024}).
\newblock


\bibitem[Li et~al\mbox{.}(2023)]%
        {li2023source}
\bibfield{author}{\bibinfo{person}{Kai Li}, \bibinfo{person}{Deep Patel}, \bibinfo{person}{Erik Kruus}, {and} \bibinfo{person}{Martin~Renqiang Min}.} \bibinfo{year}{2023}\natexlab{}.
\newblock \showarticletitle{Source-free video domain adaptation with spatial-temporal-historical consistency learning}. In \bibinfo{booktitle}{\emph{Proceedings of the IEEE/CVF Conference on Computer Vision and Pattern Recognition}}. \bibinfo{pages}{14643--14652}.
\newblock


\bibitem[Li et~al\mbox{.}(2020)]%
        {GAN1}
\bibfield{author}{\bibinfo{person}{Rui Li}, \bibinfo{person}{Qianfen Jiao}, \bibinfo{person}{Wenming Cao}, \bibinfo{person}{Hau-San Wong}, {and} \bibinfo{person}{Si Wu}.} \bibinfo{year}{2020}\natexlab{}.
\newblock \showarticletitle{Model adaptation: Unsupervised domain adaptation without source data}. In \bibinfo{booktitle}{\emph{Proceedings of the IEEE/CVF Conference on Computer Vision and Pattern Recognition}}. \bibinfo{pages}{9641--9650}.
\newblock


\bibitem[Liang et~al\mbox{.}(2020)]%
        {shot}
\bibfield{author}{\bibinfo{person}{Jian Liang}, \bibinfo{person}{Dapeng Hu}, {and} \bibinfo{person}{Jiashi Feng}.} \bibinfo{year}{2020}\natexlab{}.
\newblock \showarticletitle{Do we really need to access the source data? Source hypothesis transfer for unsupervised domain adaptation}. In \bibinfo{booktitle}{\emph{International Conference on Machine Learning}}. PMLR, \bibinfo{pages}{6028--6039}.
\newblock


\bibitem[Liang et~al\mbox{.}(2021)]%
        {SHOT++}
\bibfield{author}{\bibinfo{person}{Jian Liang}, \bibinfo{person}{Dapeng Hu}, \bibinfo{person}{Yunbo Wang}, \bibinfo{person}{Ran He}, {and} \bibinfo{person}{Jiashi Feng}.} \bibinfo{year}{2021}\natexlab{}.
\newblock \showarticletitle{Source data-absent unsupervised domain adaptation through hypothesis transfer and labeling transfer}.
\newblock \bibinfo{journal}{\emph{IEEE Transactions on Pattern Analysis and Machine Intelligence}} \bibinfo{volume}{44}, \bibinfo{number}{11} (\bibinfo{year}{2021}), \bibinfo{pages}{8602--8617}.
\newblock


\bibitem[Liu et~al\mbox{.}(2024)]%
        {liu2024graph}
\bibfield{author}{\bibinfo{person}{Chenyu Liu}, \bibinfo{person}{Xinliang Zhou}, \bibinfo{person}{Yihao Wu}, \bibinfo{person}{Ruizhi Yang}, \bibinfo{person}{Zhongruo Wang}, \bibinfo{person}{Liming Zhai}, \bibinfo{person}{Ziyu Jia}, {and} \bibinfo{person}{Yang Liu}.} \bibinfo{year}{2024}\natexlab{}.
\newblock \showarticletitle{Graph neural networks in EEG-based emotion recognition: A survey}.
\newblock \bibinfo{journal}{\emph{arXiv preprint arXiv:2402.01138}} (\bibinfo{year}{2024}).
\newblock


\bibitem[Liu and Xue(2021)]%
        {dskn}
\bibfield{author}{\bibinfo{person}{Qiao Liu} {and} \bibinfo{person}{Hui Xue}.} \bibinfo{year}{2021}\natexlab{}.
\newblock \showarticletitle{Adversarial spectral kernel matching for unsupervised time series domain adaptation}. In \bibinfo{booktitle}{\emph{Proceedings of the Thirtieth International Joint Conference on Artificial Intelligence}}. \bibinfo{pages}{2744--2750}.
\newblock


\bibitem[Long et~al\mbox{.}(2018)]%
        {CDAN}
\bibfield{author}{\bibinfo{person}{Mingsheng Long}, \bibinfo{person}{Zhangjie Cao}, \bibinfo{person}{Jianmin Wang}, {and} \bibinfo{person}{Michael~I Jordan}.} \bibinfo{year}{2018}\natexlab{}.
\newblock \showarticletitle{Conditional adversarial domain adaptation}.
\newblock \bibinfo{journal}{\emph{Advances in Neural Information Processing Systems}}  \bibinfo{volume}{31} (\bibinfo{year}{2018}).
\newblock


\bibitem[Mao et~al\mbox{.}(2024)]%
        {Entropymin1}
\bibfield{author}{\bibinfo{person}{Haitao Mao}, \bibinfo{person}{Lun Du}, \bibinfo{person}{Yujia Zheng}, \bibinfo{person}{Qiang Fu}, \bibinfo{person}{Zelin Li}, \bibinfo{person}{Xu Chen}, \bibinfo{person}{Shi Han}, {and} \bibinfo{person}{Dongmei Zhang}.} \bibinfo{year}{2024}\natexlab{}.
\newblock \showarticletitle{Source free graph unsupervised domain adaptation}. In \bibinfo{booktitle}{\emph{Proceedings of the 17th ACM International Conference on Web Search and Data Mining}}. \bibinfo{pages}{520--528}.
\newblock


\bibitem[M{\"u}ller et~al\mbox{.}(2019)]%
        {label_smoothing}
\bibfield{author}{\bibinfo{person}{Rafael M{\"u}ller}, \bibinfo{person}{Simon Kornblith}, {and} \bibinfo{person}{Geoffrey~E Hinton}.} \bibinfo{year}{2019}\natexlab{}.
\newblock \showarticletitle{When does label smoothing help?}
\newblock \bibinfo{journal}{\emph{Advances in Neural Information Processing Systems}}  \bibinfo{volume}{32} (\bibinfo{year}{2019}).
\newblock


\bibitem[Ott et~al\mbox{.}(2022)]%
        {ottUDA}
\bibfield{author}{\bibinfo{person}{Felix Ott}, \bibinfo{person}{David R{\"u}gamer}, \bibinfo{person}{Lucas Heublein}, \bibinfo{person}{Bernd Bischl}, {and} \bibinfo{person}{Christopher Mutschler}.} \bibinfo{year}{2022}\natexlab{}.
\newblock \showarticletitle{Domain adaptation for time-series classification to mitigate covariate shift}. In \bibinfo{booktitle}{\emph{Proceedings of the 30th ACM International Conference on Multimedia}}. \bibinfo{pages}{5934--5943}.
\newblock


\bibitem[Qiu et~al\mbox{.}(2021)]%
        {Select2}
\bibfield{author}{\bibinfo{person}{Zhen Qiu}, \bibinfo{person}{Yifan Zhang}, \bibinfo{person}{Hongbin Lin}, \bibinfo{person}{Shuaicheng Niu}, \bibinfo{person}{Yanxia Liu}, \bibinfo{person}{Qing Du}, {and} \bibinfo{person}{Mingkui Tan}.} \bibinfo{year}{2021}\natexlab{}.
\newblock \showarticletitle{Source-free domain adaptation via avatar prototype generation and adaptation}. In \bibinfo{booktitle}{\emph{International Joint Conference on Artificial Intelligence}}.
\newblock


\bibitem[Ragab et~al\mbox{.}(2022)]%
        {SLARDA}
\bibfield{author}{\bibinfo{person}{Mohamed Ragab}, \bibinfo{person}{Emadeldeen Eldele}, \bibinfo{person}{Zhenghua Chen}, \bibinfo{person}{Min Wu}, \bibinfo{person}{Chee-Keong Kwoh}, {and} \bibinfo{person}{Xiaoli Li}.} \bibinfo{year}{2022}\natexlab{}.
\newblock \showarticletitle{Self-supervised autoregressive domain adaptation for time series data}.
\newblock \bibinfo{journal}{\emph{IEEE Transactions on Neural Networks and Learning Systems}} \bibinfo{volume}{35}, \bibinfo{number}{1} (\bibinfo{year}{2022}), \bibinfo{pages}{1341--1351}.
\newblock


\bibitem[Ragab et~al\mbox{.}(2023a)]%
        {adatime}
\bibfield{author}{\bibinfo{person}{Mohamed Ragab}, \bibinfo{person}{Emadeldeen Eldele}, \bibinfo{person}{Wee~Ling Tan}, \bibinfo{person}{Chuan-Sheng Foo}, \bibinfo{person}{Zhenghua Chen}, \bibinfo{person}{Min Wu}, \bibinfo{person}{Chee-Keong Kwoh}, {and} \bibinfo{person}{Xiaoli Li}.} \bibinfo{year}{2023}\natexlab{a}.
\newblock \showarticletitle{Adatime: A benchmarking suite for domain adaptation on time series data}.
\newblock \bibinfo{journal}{\emph{ACM Transactions on Knowledge Discovery from Data}} \bibinfo{volume}{17}, \bibinfo{number}{8} (\bibinfo{year}{2023}), \bibinfo{pages}{1--18}.
\newblock


\bibitem[Ragab et~al\mbox{.}(2023b)]%
        {MAPU}
\bibfield{author}{\bibinfo{person}{Mohamed Ragab}, \bibinfo{person}{Emadeldeen Eldele}, \bibinfo{person}{Min Wu}, \bibinfo{person}{Chuan-Sheng Foo}, \bibinfo{person}{Xiaoli Li}, {and} \bibinfo{person}{Zhenghua Chen}.} \bibinfo{year}{2023}\natexlab{b}.
\newblock \showarticletitle{Source-free domain adaptation with temporal imputation for time series data}. In \bibinfo{booktitle}{\emph{Proceedings of the 29th ACM SIGKDD Conference on Knowledge Discovery and Data Mining}}. \bibinfo{pages}{1989--1998}.
\newblock


\bibitem[Ragab et~al\mbox{.}(2024)]%
        {E_MAPU}
\bibfield{author}{\bibinfo{person}{Mohamed Ragab}, \bibinfo{person}{Peiliang Gong}, \bibinfo{person}{Emadeldeen Eldele}, \bibinfo{person}{Wenyu Zhang}, \bibinfo{person}{Min Wu}, \bibinfo{person}{Chuan-Sheng Foo}, \bibinfo{person}{Daoqiang Zhang}, \bibinfo{person}{Xiaoli Li}, {and} \bibinfo{person}{Zhenghua Chen}.} \bibinfo{year}{2024}\natexlab{}.
\newblock \showarticletitle{Evidentially calibrated source-free time-series domain adaptation with temporal imputation}.
\newblock \bibinfo{journal}{\emph{arXiv e-prints}} (\bibinfo{year}{2024}), \bibinfo{pages}{arXiv--2406}.
\newblock


\bibitem[{Rahman} et~al\mbox{.}(2020)]%
        {MMDA}
\bibfield{author}{\bibinfo{person}{Mohammad~Mahfujur {Rahman}}, \bibinfo{person}{Clinton {Fookes}}, \bibinfo{person}{Mahsa {Baktashmotlagh}}, {and} \bibinfo{person}{Sridha {Sridharan}}.} \bibinfo{year}{2020}\natexlab{}.
\newblock \showarticletitle{On minimum discrepancy estimation for deep domain adaptation}.
\newblock \bibinfo{journal}{\emph{Domain Adaptation for Visual Understanding}} (\bibinfo{year}{2020}).
\newblock


\bibitem[Sun et~al\mbox{.}(2017)]%
        {deep_coral}
\bibfield{author}{\bibinfo{person}{Baochen Sun}, \bibinfo{person}{Jiashi Feng}, {and} \bibinfo{person}{Kate Saenko}.} \bibinfo{year}{2017}\natexlab{}.
\newblock \showarticletitle{Correlation alignment for unsupervised domain adaptation}.
\newblock In \bibinfo{booktitle}{\emph{Domain Adaptation in Computer Vision Applications}}. \bibinfo{publisher}{Springer}, \bibinfo{pages}{153--171}.
\newblock


\bibitem[Tzeng et~al\mbox{.}(2014)]%
        {ddc}
\bibfield{author}{\bibinfo{person}{Eric Tzeng}, \bibinfo{person}{Judy Hoffman}, \bibinfo{person}{Ning Zhang}, \bibinfo{person}{Kate Saenko}, {and} \bibinfo{person}{Trevor Darrell}.} \bibinfo{year}{2014}\natexlab{}.
\newblock \showarticletitle{Deep domain confusion: Maximizing for domain invariance}.
\newblock \bibinfo{journal}{\emph{arXiv preprint arXiv:1412.3474}} (\bibinfo{year}{2014}).
\newblock


\bibitem[Van~der Maaten and Hinton(2008)]%
        {tSNE}
\bibfield{author}{\bibinfo{person}{Laurens Van~der Maaten} {and} \bibinfo{person}{Geoffrey Hinton}.} \bibinfo{year}{2008}\natexlab{}.
\newblock \showarticletitle{Visualizing data using t-SNE.}
\newblock \bibinfo{journal}{\emph{Journal of Machine Learning Research}} \bibinfo{volume}{9}, \bibinfo{number}{11} (\bibinfo{year}{2008}).
\newblock


\bibitem[Wang et~al\mbox{.}(2024a)]%
        {pond}
\bibfield{author}{\bibinfo{person}{Junxiang Wang}, \bibinfo{person}{Guangji Bai}, \bibinfo{person}{Wei Cheng}, \bibinfo{person}{Zhengzhang Chen}, \bibinfo{person}{Liang Zhao}, {and} \bibinfo{person}{Haifeng Chen}.} \bibinfo{year}{2024}\natexlab{a}.
\newblock \showarticletitle{POND: Multi-source time series domain adaptation with information-aware prompt tuning}. In \bibinfo{booktitle}{\emph{Proceedings of the 30th ACM SIGKDD Conference on Knowledge Discovery and Data Mining}} (Barcelona, Spain) \emph{(\bibinfo{series}{KDD '24})}. \bibinfo{address}{New York, NY, USA}, \bibinfo{pages}{3140--3151}.
\newblock


\bibitem[Wang et~al\mbox{.}(2024b)]%
        {wang2024temporal}
\bibfield{author}{\bibinfo{person}{Yucheng Wang}, \bibinfo{person}{Peiliang Gong}, \bibinfo{person}{Min Wu}, \bibinfo{person}{Felix Ott}, \bibinfo{person}{Xiaoli Li}, \bibinfo{person}{Lihua Xie}, {and} \bibinfo{person}{Zhenghua Chen}.} \bibinfo{year}{2024}\natexlab{b}.
\newblock \showarticletitle{Temporal source recovery for time-series source-free unsupervised domain adaptation}.
\newblock \bibinfo{journal}{\emph{arXiv preprint arXiv:2409.19635}} (\bibinfo{year}{2024}).
\newblock


\bibitem[Wang et~al\mbox{.}(2025)]%
        {wang2025survey}
\bibfield{author}{\bibinfo{person}{Yucheng Wang}, \bibinfo{person}{Min Wu}, \bibinfo{person}{Xiaoli Li}, \bibinfo{person}{Lihua Xie}, {and} \bibinfo{person}{Zhenghua Chen}.} \bibinfo{year}{2025}\natexlab{}.
\newblock \showarticletitle{A survey on graph neural networks for remaining useful life prediction: Methodologies, evaluation and future trends}.
\newblock \bibinfo{journal}{\emph{Mechanical Systems and Signal Processing}}  \bibinfo{volume}{229} (\bibinfo{year}{2025}), \bibinfo{pages}{112449}.
\newblock


\bibitem[Wang et~al\mbox{.}(2023)]%
        {SEA}
\bibfield{author}{\bibinfo{person}{Yucheng Wang}, \bibinfo{person}{Yuecong Xu}, \bibinfo{person}{Jianfei Yang}, \bibinfo{person}{Zhenghua Chen}, \bibinfo{person}{Min Wu}, \bibinfo{person}{Xiaoli Li}, {and} \bibinfo{person}{Lihua Xie}.} \bibinfo{year}{2023}\natexlab{}.
\newblock \showarticletitle{Sensor alignment for multivariate time-series unsupervised domain adaptation}. In \bibinfo{booktitle}{\emph{Proceedings of the AAAI Conference on Artificial Intelligence}}, Vol.~\bibinfo{volume}{37}. \bibinfo{pages}{10253--10261}.
\newblock


\bibitem[Wang et~al\mbox{.}(2024c)]%
        {SEA++}
\bibfield{author}{\bibinfo{person}{Yucheng Wang}, \bibinfo{person}{Yuecong Xu}, \bibinfo{person}{Jianfei Yang}, \bibinfo{person}{Min Wu}, \bibinfo{person}{Xiaoli Li}, \bibinfo{person}{Lihua Xie}, {and} \bibinfo{person}{Zhenghua Chen}.} \bibinfo{year}{2024}\natexlab{c}.
\newblock \showarticletitle{SEA++: Multi-graph-based higher-order sensor alignment for multivariate time-series unsupervised domain adaptation}.
\newblock \bibinfo{journal}{\emph{IEEE Transactions on Pattern Analysis and Machine Intelligence}} \bibinfo{volume}{46}, \bibinfo{number}{12} (\bibinfo{year}{2024}), \bibinfo{pages}{10781--10796}.
\newblock


\bibitem[Wilson et~al\mbox{.}(2020)]%
        {codats}
\bibfield{author}{\bibinfo{person}{Garrett Wilson}, \bibinfo{person}{Janardhan~Rao Doppa}, {and} \bibinfo{person}{Diane~J Cook}.} \bibinfo{year}{2020}\natexlab{}.
\newblock \showarticletitle{Multi-source deep domain adaptation with weak supervision for time-series sensor data}. In \bibinfo{booktitle}{\emph{Proceedings of the 26th ACM SIGKDD International Conference on Knowledge Discovery \& Data Mining}}. \bibinfo{pages}{1768--1778}.
\newblock


\bibitem[Wilson et~al\mbox{.}(2023)]%
        {CALDA}
\bibfield{author}{\bibinfo{person}{Garrett Wilson}, \bibinfo{person}{Janardhan~Rao Doppa}, {and} \bibinfo{person}{Diane~J Cook}.} \bibinfo{year}{2023}\natexlab{}.
\newblock \showarticletitle{CALDA: Improving multi-source time series domain adaptation with contrastive adversarial learning}.
\newblock \bibinfo{journal}{\emph{IEEE Transactions on Pattern Analysis and Machine Intelligence}} \bibinfo{volume}{45}, \bibinfo{number}{12} (\bibinfo{year}{2023}), \bibinfo{pages}{14208--14221}.
\newblock


\bibitem[Wu et~al\mbox{.}(2020)]%
        {MTGNN}
\bibfield{author}{\bibinfo{person}{Zonghan Wu}, \bibinfo{person}{Shirui Pan}, \bibinfo{person}{Guodong Long}, \bibinfo{person}{Jing Jiang}, \bibinfo{person}{Xiaojun Chang}, {and} \bibinfo{person}{Chengqi Zhang}.} \bibinfo{year}{2020}\natexlab{}.
\newblock \showarticletitle{Connecting the dots: Multivariate time series forecasting with graph neural networks}. In \bibinfo{booktitle}{\emph{Proceedings of the 26th ACM SIGKDD International Conference on Knowledge Discovery \& Data Mining}} (Virtual Event, CA, USA) \emph{(\bibinfo{series}{KDD '20})}. \bibinfo{address}{New York, NY, USA}, \bibinfo{pages}{753–763}.
\newblock


\bibitem[Xie et~al\mbox{.}(2022)]%
        {Pseudolabel1}
\bibfield{author}{\bibinfo{person}{Binhui Xie}, \bibinfo{person}{Longhui Yuan}, \bibinfo{person}{Shuang Li}, \bibinfo{person}{Chi~Harold Liu}, \bibinfo{person}{Xinjing Cheng}, {and} \bibinfo{person}{Guoren Wang}.} \bibinfo{year}{2022}\natexlab{}.
\newblock \showarticletitle{Active learning for domain adaptation: An energy-based approach}. In \bibinfo{booktitle}{\emph{Proceedings of the AAAI Conference on Artificial Intelligence}}, Vol.~\bibinfo{volume}{36}. \bibinfo{pages}{8708--8716}.
\newblock


\bibitem[Xu et~al\mbox{.}(2022)]%
        {xu2022source}
\bibfield{author}{\bibinfo{person}{Yuecong Xu}, \bibinfo{person}{Jianfei Yang}, \bibinfo{person}{Haozhi Cao}, \bibinfo{person}{Keyu Wu}, \bibinfo{person}{Min Wu}, {and} \bibinfo{person}{Zhenghua Chen}.} \bibinfo{year}{2022}\natexlab{}.
\newblock \showarticletitle{Source-free video domain adaptation by learning temporal consistency for action recognition}. In \bibinfo{booktitle}{\emph{European Conference on Computer Vision}}. Springer, \bibinfo{pages}{147--164}.
\newblock


\bibitem[Yang et~al\mbox{.}(2021)]%
        {nrc}
\bibfield{author}{\bibinfo{person}{Shiqi Yang}, \bibinfo{person}{Joost van~de Weijer}, \bibinfo{person}{Luis Herranz}, \bibinfo{person}{Shangling Jui}, {et~al\mbox{.}}} \bibinfo{year}{2021}\natexlab{}.
\newblock \showarticletitle{Exploiting the intrinsic neighborhood structure for source-free domain adaptation}.
\newblock \bibinfo{journal}{\emph{Advances in Neural Information Processing Systems}}  \bibinfo{volume}{34} (\bibinfo{year}{2021}), \bibinfo{pages}{29393--29405}.
\newblock


\bibitem[Yang et~al\mbox{.}(2022)]%
        {aad}
\bibfield{author}{\bibinfo{person}{Shiqi Yang}, \bibinfo{person}{Yaxing Wang}, \bibinfo{person}{Kai Wang}, \bibinfo{person}{Shangling Jui}, {et~al\mbox{.}}} \bibinfo{year}{2022}\natexlab{}.
\newblock \showarticletitle{Attracting and dispersing: A simple approach for source-free domain adaptation}. In \bibinfo{booktitle}{\emph{Advances in Neural Information Processing Systems}}.
\newblock


\bibitem[Zhang et~al\mbox{.}(2023a)]%
        {zhang2023temporal}
\bibfield{author}{\bibinfo{person}{Junru Zhang}, \bibinfo{person}{Lang Feng}, \bibinfo{person}{Yang He}, \bibinfo{person}{Yuhan Wu}, {and} \bibinfo{person}{Yabo Dong}.} \bibinfo{year}{2023}\natexlab{a}.
\newblock \showarticletitle{Temporal convolutional explorer helps understand 1d-cnn's learning behavior in time series classification from frequency domain}. In \bibinfo{booktitle}{\emph{Proceedings of the 32nd ACM International Conference on Information and Knowledge Management}}. \bibinfo{pages}{3351--3360}.
\newblock


\bibitem[Zhang et~al\mbox{.}(2024)]%
        {zhang2024diverse}
\bibfield{author}{\bibinfo{person}{Junru Zhang}, \bibinfo{person}{Lang Feng}, \bibinfo{person}{Zhidan Liu}, \bibinfo{person}{Yuhan Wu}, \bibinfo{person}{Yang He}, \bibinfo{person}{Yabo Dong}, {and} \bibinfo{person}{Duanqing Xu}.} \bibinfo{year}{2024}\natexlab{}.
\newblock \showarticletitle{Diverse intra-and inter-domain activity style fusion for cross-person generalization in activity recognition}. In \bibinfo{booktitle}{\emph{Proceedings of the 30th ACM SIGKDD Conference on Knowledge Discovery and Data Mining}}. \bibinfo{pages}{4213--4222}.
\newblock


\bibitem[Zhang et~al\mbox{.}(2023b)]%
        {zhang2023adacket}
\bibfield{author}{\bibinfo{person}{Junru Zhang}, \bibinfo{person}{Lang Feng}, \bibinfo{person}{Haowen Zhang}, \bibinfo{person}{Yuhan Wu}, {and} \bibinfo{person}{Yabo Dong}.} \bibinfo{year}{2023}\natexlab{b}.
\newblock \showarticletitle{Adacket: Adaptive convolutional kernel transform for multivariate time series classification}. In \bibinfo{booktitle}{\emph{Joint European Conference on Machine Learning and Knowledge Discovery in Databases}}. Springer, \bibinfo{pages}{189--204}.
\newblock


\bibitem[Zhang et~al\mbox{.}(2022)]%
        {DAC}
\bibfield{author}{\bibinfo{person}{Ziyi Zhang}, \bibinfo{person}{Weikai Chen}, \bibinfo{person}{Hui Cheng}, \bibinfo{person}{Zhen Li}, \bibinfo{person}{Siyuan Li}, \bibinfo{person}{Liang Lin}, {and} \bibinfo{person}{Guanbin Li}.} \bibinfo{year}{2022}\natexlab{}.
\newblock \showarticletitle{Divide and contrast: Source-free domain adaptation via adaptive contrastive learning}.
\newblock \bibinfo{journal}{\emph{Advances in Neural Information Processing Systems}}  \bibinfo{volume}{35} (\bibinfo{year}{2022}), \bibinfo{pages}{5137--5149}.
\newblock


\end{thebibliography}
